\documentclass[letterpaper]{article} % DO NOT CHANGE THIS
\usepackage{aaai2026}  % DO NOT CHANGE THIS
\usepackage{times}  % DO NOT CHANGE THIS
\usepackage{helvet}  % DO NOT CHANGE THIS
\usepackage{courier}  % DO NOT CHANGE THIS
\usepackage[hyphens]{url}  % DO NOT CHANGE THIS
\usepackage{graphicx} % DO NOT CHANGE THIS
\usepackage{natbib}  % DO NOT CHANGE THIS AND DO NOT ADD ANY OPTIONS TO IT
\usepackage{caption} % DO NOT CHANGE THIS AND DO NOT ADD ANY OPTIONS TO IT
\usepackage{algorithm}
\usepackage{algorithmic}
\usepackage{fontawesome5}

\usepackage{amsmath}
\usepackage{multirow}
\usepackage{booktabs}
\usepackage{tabularray}
\usepackage{pifont}

\usepackage{newfloat}
\usepackage{listings}
\DeclareCaptionStyle{ruled}{labelfont=normalfont,labelsep=colon,strut=off} % DO NOT CHANGE THIS
\floatstyle{ruled}
\newfloat{listing}{tb}{lst}{}
\floatname{listing}{Listing}
\title{
 Unified Start, Personalized End: Progressive Pruning for Efficient 3D Medical Image Segmentation
}
\author {
    Linhao Li\textsuperscript{\rm 1} \equalcontrib,
    Yiwen Ye\textsuperscript{\rm 1} \equalcontrib,
    Ziyang Chen\textsuperscript{\rm 1},
    Yong Xia\textsuperscript{\rm 1 \thanks{Corresponding author: Y. Xia.}} 
}
\affiliations {
    \textsuperscript{\rm 1}National Engineering Laboratory for Integrated Aero-Space-Ground-Ocean Big Data Application Technology, School of
 Computer Science and Engineering, Northwestern Polytechnical University, China\\
    {lilinhao, ywye, zychen}@mail.nwpu.edu.cn, yxia@nwpu.edu.cn
}

\begin{document}

\maketitle

\begin{figure*}[t]
\centerline{\includegraphics[width=0.95\linewidth]{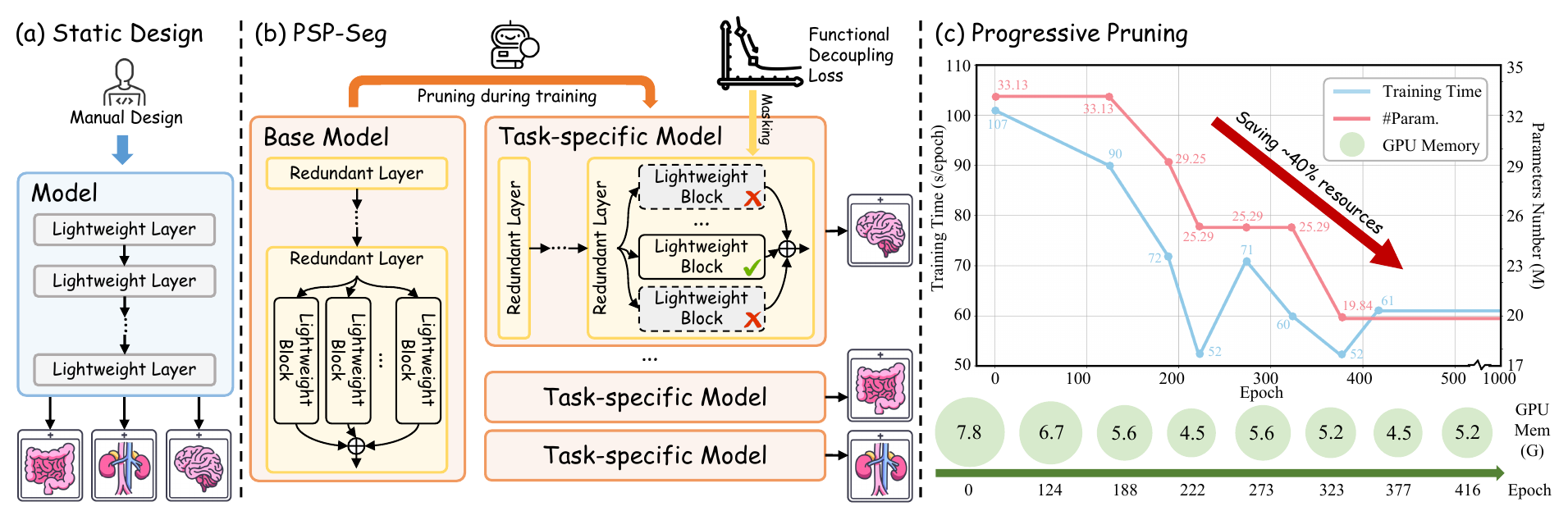}}
\caption{Comparison of static models and our PSP-Seg. (a) Static Model: Pre-design model structure before training, and is used to adapt to various tasks, \textit{i.e.}, different tasks share the same model structure. (b) PSP-Seg: Pre-design a redundant model, and progressive pruning during the training based on functional decoupling losses (FD Loss), \textit{i.e.}, each task has a unique model. (c) Progressive pruning: The redundant model is progressively pruned into a task-specific model, saving approximately 40\% of the resources.  We adopt a mask-then-prune strategy to avoid over-pruning, therefore, the broken line is not always downward.}
\label{fig:intro}
\end{figure*}

\begin{abstract}
3D medical image segmentation often faces heavy resource and time consumption, limiting its scalability and rapid deployment in clinical environments. Existing efficient segmentation models are typically static and manually designed prior to training, which restricts their adaptability across diverse tasks and makes it difficult to balance performance with resource efficiency. In this paper, we propose PSP-Seg, a progressive pruning framework that enables dynamic and efficient 3D segmentation. PSP-Seg begins with a redundant model and iteratively prunes redundant modules through a combination of block-wise pruning and a functional decoupling loss. We evaluate PSP-Seg on five public datasets, benchmarking it against seven state-of-the-art models and six efficient segmentation models. Results demonstrate that the lightweight variant, PSP-Seg-S, achieves performance on par with nnU-Net while reducing GPU memory usage by 42–45\%, training time by 29–48\%, and parameter number by 83–87\% across all datasets. These findings underscore PSP-Seg’s potential as a cost-effective yet high-performing alternative for widespread clinical application. Code and weights will be available once accepted.

\end{abstract}

\begin{links}
    \link{Code}{https://github.com/leoLilh/PSP-Seg}
\end{links}

\section{Introduction}
% 3D分割重要 -〉 现有的方法有\textbf{很好的效果}，但是资源需求大 -〉因此轻量化模型被研究者重视 -〉现有的轻量化方法以构建轻量化模型为主 -〉这些方法是静态的，忽略了不同任务难度不同，需要的参数量也应不一样 -〉因为我们引入在训练中逐步剪枝的策略
The accurate extraction of organs, tissues, and tumors from volumetric medical data is fundamental to modern healthcare, supporting precise diagnoses and personalized treatment planning \cite{azad2024medical,rayed2024deep}. Recent advances in deep learning have enabled the development of powerful 3D segmentation models, such as nnU-Net  \cite{isensee2021nnu} and STU-Net \cite{huang2023stu}, which have shown outstanding performance across a variety of segmentation tasks. However, the high performance of these models often comes at the cost of significant computational and resource requirements. They typically require long training times, large GPU memory allocations, and substantial storage for their millions of parameters. For example, training U-Mamba with a patch size of $128\times128\times128$ and a batch size of 2 consumes over 36 hours and 12 GB of GPU memory, with 70M parameters. Such resource-heavy models hinder rapid product iteration and limit deployment in clinical settings with constrained computational infrastructure. This challenge has prompted growing interest in lightweight or efficiency-aware segmentation approaches that aim to maintain strong performance while significantly reducing resource overhead.
To address this, existing efficient segmentation methods primarily focus on the design of compact model architectures. These include module-level optimizations \cite{chen2019dmfnet, sadegheih2024lhunet, pang2024slimunetr, yadav2025mlru++}, which replace backbone components with more efficient alternatives, and architecture-level strategies \cite{perera2024segformer3d, rahman2025effidec3d}, which aim to eliminate structural redundancies. While these methods successfully reduce FLOPs and parameter numbers \cite{pang2024slimunetr}, they are typically static and pre-designed before training on the target data. As illustrated in Fig. \ref{fig:intro}(a), this one-size-fits-all design overlooks the diversity and task-specific characteristics inherent to medical image segmentation. We argue that \textbf{a static model cannot universally fit all tasks\footnote{A model is considered ‘suitable’ for a task if, under certain resource constraints, it can achieve strong performance with minimal parameter or structural redundancy.}, and it is impractical to determine the optimal model configuration in advance.}

Motivated by these limitations, we propose \textbf{P}rogre\textbf{S}sive \textbf{P}runing Segmentation framework (PSP-Seg), a framework for efficient 3D medical image segmentation. Unlike static models, PSP-Seg begins with a redundant model\footnote{We define “redundant” as using all available GPU memory to construct a highly expressive model.} and progressively prunes away unnecessary components during training. This progressive approach allows the model to dynamically adapt its structure to the specific task at hand, yielding a compact, task-specific architecture, as shown in Fig. \ref{fig:intro}(c). 
Specifically, 
PSP-Seg introduces a parallel redundant module to initialize the model with sufficient capacity for task learning. A task-oriented progressive pruning strategy is then applied, based on a masked-then-pruned paradigm guided by functional decoupling losses. These losses help identify and decouple redundant parts of the model, steering the pruning process intelligently. Our approach continuously monitors the decoupling loss; once it converges, the loss is compared to the historical best. If improvement is observed, block-wise pruning is applied to mask redundant components. To safeguard against suboptimal pruning, the model evaluates whether performance improves post-masking. If so, the layers are permanently pruned; if not, the model restores them and applies a finer pruning step. This mechanism supports gradual convergence toward an efficient architecture without compromising task performance. 
As visualized in Fig. \ref{fig:intro}(c), PSP-Seg gradually reduces training time, parameter number, and GPU memory consumption over the course of training, achieving approximately a 40\% reduction in resource usage. We evaluate PSP-Seg on five challenging 3D medical image segmentation datasets (spanning CT and MRI modalities), and compare it with state-of-the-art segmentation methods, including CNN-based \cite{isensee2021nnu, huang2023stu}, Transformer-based \cite{zhou2021nnformer, he2023swinunetr}, Mamba-based \cite{ma2024umamba}, hybrid \cite{xie2021cotr}, and efficient models \cite{chen2019dmfnet, perera2024segformer3d, sadegheih2024lhunet, pang2024slimunetr, rahman2025effidec3d, yadav2025mlru++}. 
Since commonly used metrics like GFLOPs may not fully capture real-world efficiency during training and inference \cite{ma2018shufflenetv2, pang2024slimunetr}, we adopt more practical metrics: time and GPU memory usage of training and inference, and parameter number. Our results show that the lightweight version, PSP-Seg-S, achieves performance comparable to nnU-Net while reducing GPU memory usage by 42-45\%, training time by 29-48\%, and parameter number by 83-87\% across five datasets, demonstrating strong efficiency–performance trade-offs.
Our main contributions are:
\begin{itemize}
\item  We propose PSP-Seg, a task-oriented 3D medical image segmentation framework that achieves efficient segmentation through progressive pruning, balancing performance and resource requirement.
\item We design a redundant architecture composed of parallel redundant modules, and prune it progressively using the functional decoupling loss that intelligently guides the pruning process.
\item We employ practical metrics, training/inference time, GPU memory usage, and parameter number, to rigorously evaluate and compare model efficiency.
\item We conduct extensive experiments on five challenging 3D segmentation datasets, comparing PSP-Seg with strong baselines, including nnU-Net, STU-Net, SegFormer3D, and EffiDec3D. PSP-Seg achieves superior efficiency while maintaining high segmentation performance.
\end{itemize}

\begin{figure*}[t]
\centerline{\includegraphics[width=0.95\linewidth]{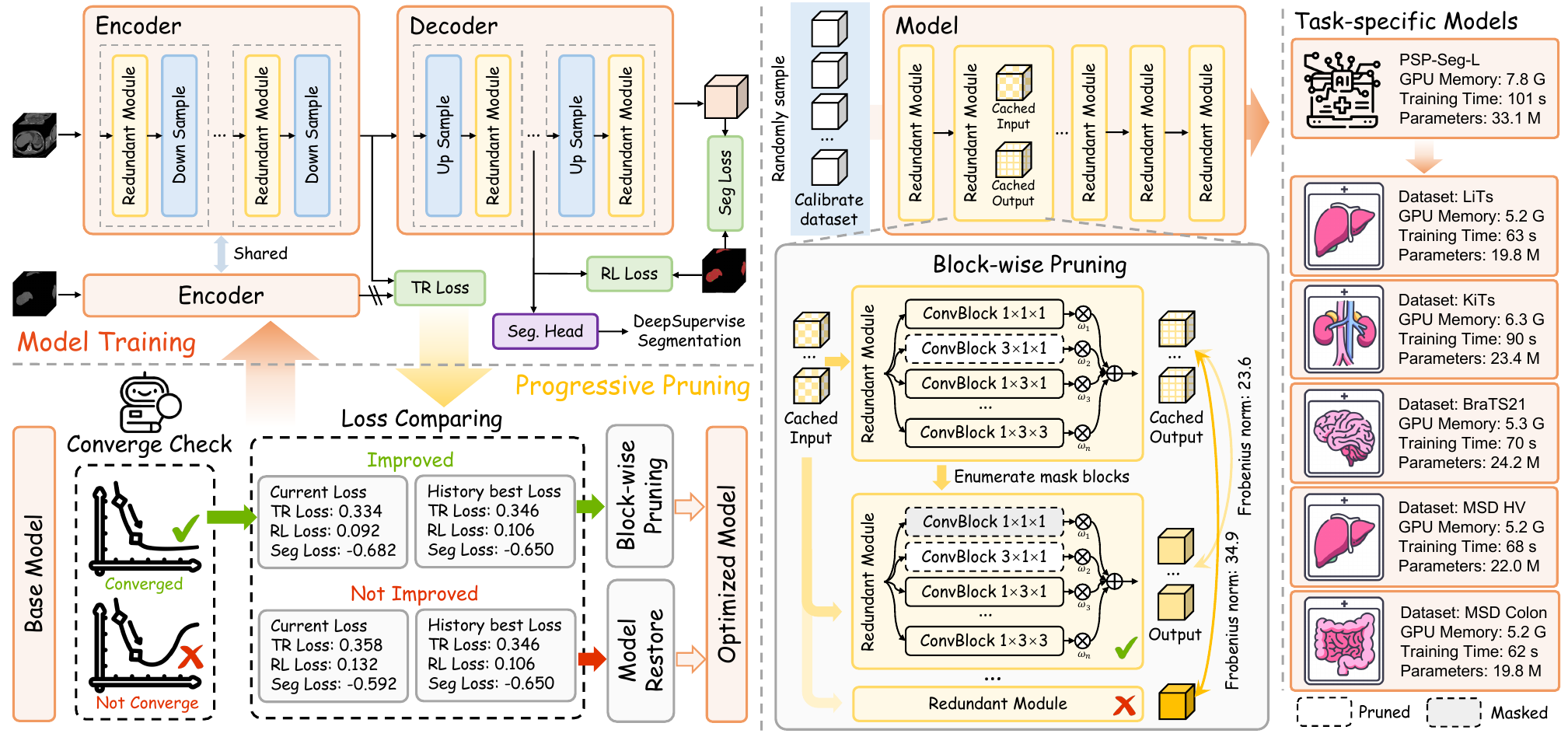}}
\caption{Overview of the PSP-Seg Framework. Each training epoch in PSP-Seg consists of two core stages: normal training and progressive pruning. During normal training, we begin with a redundant base model and optimize it using the functional decoupling loss (FD Loss) to enable expressive task learning. In progressive pruning, we apply a masked-then-pruned strategy guided by the FD Loss to gradually remove redundant components from the model. This iterative process refines the architecture toward task-specific efficiency. The final compact model generated by PSP-Seg is shown at the right.}
\label{fig:overview}
\end{figure*}

\section{Related Work}
\subsection{Efficient Medical Image Segmentation}
The evolution of 3D medical image segmentation has witnessed a progression from CNN-based models \cite{ronneberger2015unet, isensee2021nnu, huang2023stu, roy2023mednext}, to Transformer-based architectures \cite{hatamizadeh2021swin, zhou2021nnformer, gao2022medformer, he2023swinunetr}, and more recently, to Mamba-based models \cite{ma2024umamba}. Hybrid models that combine CNNs and Transformers have also emerged as strong performers \cite{chen2021transunet, xie2021cotr, sadegheih2024lhunet}. While these models achieve state-of-the-art segmentation performance, heir high computational and memory demands significantly limit their applicability in resource-constrained environments and impede model iteration in clinical workflows.
To mitigate these challenges, recent research has focused on developing efficient segmentation approaches that reduce resource consumption while preserving competitive performance. These efforts can be broadly categorized into two strategies: module efficiency \cite{chen2019dmfnet, yadav2025mlru++, sadegheih2024lhunet, pang2024slimunetr} and architecture efficiency \cite{perera2024segformer3d, rahman2025effidec3d}. 
\textbf{Module efficiency} aims to replace high-cost components in model backbones with more lightweight alternatives. This includes approaches utilizing sparse convolutional layers \cite{chen2019dmfnet} and efficient attention mechanisms \cite{yadav2025mlru++, pang2024slimunetr}. For example, MLRU++ \cite{yadav2025mlru++} improves computational efficiency by replacing traditional MLPs and full convolutions with pointwise and depthwise convolutions. Slim UNETR  \cite{pang2024slimunetr} introduces hierarchical downsampling and locally-grouped multi-head self-attention, while leveraging depthwise convolutions and bottleneck blocks to construct lightweight CNN modules.
\textbf{Architecture efficiency}, in contrast, targets global structural redundancy. These methods observe that different components of a network contribute unequally to performance. For instance, SegFormer3D  \cite{perera2024segformer3d} uses a prune-MLP decoder to aggregate multi-scale features efficiently, and EffiDec3D  \cite{rahman2025effidec3d} reduces computation by removing shallow decoder layers and reducing channel dimensions.

However, efficiency inherently involves a trade-off between performance and resource usage, one that cannot be properly assessed without training data. Static models, pre-defined prior to training, often fall short in adapting to the nuanced demands of different tasks. In contrast, we argue that dynamic models, which evolve during training, are better suited to achieving an optimal balance between efficiency and performance. To this end, our proposed PSP-Seg framework adopts a progressive pruning paradigm that adaptively learns compact, task-specific architectures from an initially redundant model.

\subsection{Model Pruning}
Model pruning is a well-established technique for compressing deep models by eliminating redundant modules or connections while maintaining competitive accuracy \cite{cheng2024prunesurvey}. It is particularly valuable for deploying large models in environments with limited computational resources and for reducing operational costs. Pruning strategies can be broadly divided into two categories: pruning after training and pruning during training. 
\textbf{Pruning after training} \cite{frantar2023sparsegpt, sun2023wanda, men2024shortgpt, lu2024not} removes unimportant weights or units from a fully trained model. These methods often rely on scoring metrics to evaluate each parameter’s contribution. For example, ShortGPT \cite{men2024shortgpt} introduces the Block Influence metric to assess a layer's functional importance by measuring its effect on hidden state representations. Although these approaches can achieve high compression rates, they typically require prior knowledge or heuristics to determine pruning ratios and may necessitate multiple re-training cycles for optimal results.
\textbf{Pruning during training} \cite{li2019fcf, liu2021freetickets, he2019fpgm, cho2023pdp} integrates pruning into the training process, allowing simultaneous optimization of model weights and sparsity patterns. These methods commonly use thresholds or soft masks to guide the pruning process. For instance, Chen \textit{et al.} \cite{chen2022task} compute a proficiency score for each module and prune those falling below a predefined threshold. While these approaches offer flexibility and adaptability, they often lack mechanisms to recover from over-pruning, that is, mistakenly removing components that may later prove important.
Our proposed PSP-Seg framework aligns with the ‘Pruning during training' paradigm but introduces a more adaptive and robust strategy. PSP-Seg performs block-wise progressive pruning to dynamically remove redundant module. Crucially, it monitors the pruning process in real time and can restore previously masked components if their removal negatively impacts performance. This reversible mechanism enhances both robustness and task adaptability, enabling PSP-Seg to produce compact yet performant models across diverse medical segmentation tasks.

\begin{figure}[t]
\centerline{\includegraphics[width=0.95\linewidth]{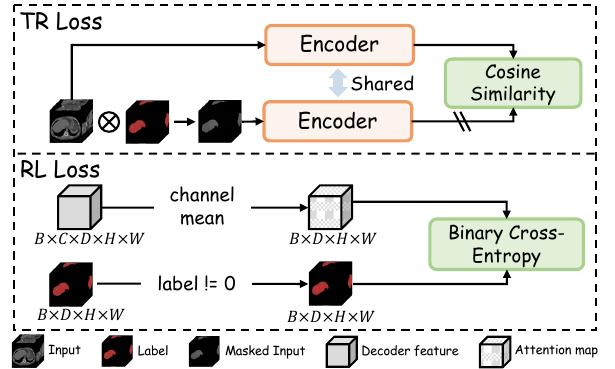}}
\caption{Illustration of the proposed functional decoupling loss, consisting of TR Loss and RL Loss. TR Loss guides the encoder to learn task-relevant features. RL Loss assists the decoder in accurately identifying and delineating the target region.}
\label{fig:loss}
\end{figure}

\section{Methodology}
\subsection{Overview}
PSP-Seg adopts a redundant-to-compact strategy: it begins with a highly redundant base model and progressively prunes it into a compact, task-specific model during training, as illustrated in Fig. \ref{fig:overview}. Each training epoch comprises two stages: (1) normal training following nnU-Net V2 \cite{isensee2021nnu}, and (2) progressive pruning.  In the first phase, the model learns knowledge from data; in the second, it prunes structurally redundant modules. Details of each component are outlined below.

\subsubsection{Redundant Base Model}
To provide sufficient capacity at initialization, PSP-Seg constructs a U-shaped redundant base model using a design called the Parallel Redundant Module (PRM). Each PRM comprises multiple parallel convolutional efficient blocks (EB) with varying kernel sizes to capture multiscale spatial context.
To ensure computational efficiency, each efficient block employs a bottleneck architecture. Specifically, a $1 \times 1 \times 1$ convolution reduces the number of channels (squeeze ratio = 0.5), followed by Instance Normalization \cite{ulyanov2016insnorm} and LeakyReLU \cite{maas2013leakyrelu}. A $k_1 \times k_2 \times k_3$ convolution then restores the original number of channels. Given an input $x$, a efficient block is defined as:
\begin{equation}
    EB(x) = C_{k_1 \times k_2 \times k_3}(LReLU(IN(C_{1\times1\times1}(x)))), \label{eq:CB}
\end{equation}
where $C_{k_1 \times k_2 \times k_3}(\cdot)$ denotes a convolution with kernel size $k_1 \times k_2 \times k_3$, where $k_1, k_2, k_3 \in {1, 3}$. The final PRM output is obtained via a learnable weighted sum of all branches, along with a residual connection:
\begin{equation}
    PRM(x)=x+\sum_{i=0}^{n} w_{i}EB_i(x), \label{eq:PRM}
\end{equation}
where $n$ is the number of parallel blocks, and $w_i$ denotes the learnable importance weight for the $i$-th block.

\subsection{Normal Training}
During normal training, PSP-Seg learns segmentation knowledge using paired images and ground truths (GTs). This phase adopts a multi-scale supervision strategy similar to nnU-Net V2, combining Dice and cross-entropy losses for segmentation. In addition, we introduce a functional decoupling (FD) loss, consisting of Target Representation (TR) Loss and Region Localization (RL) Loss, to guide the encoder and decoder toward learning complementary roles and promoting structural compactness.

\subsubsection{Target Representation Loss}
The TR loss is applied at the encoder's output to encourage the extraction of task-relevant semantic features. It uses two inputs: the original image and a GT-masked version that retains only foreground content. Both are passed through a shared encoder, yielding features $x_{feat}$ and $target_{feat}$. Their similarity is encouraged using cosine similarity:
\begin{equation}
L_{tr} = 1 - \text{sim}(\text{Enc}(x), |\text{Enc}(target)|), \label{eq:losstr}
\end{equation}
where $\text{sim}(\cdot,\cdot)$ denotes cosine similarity, $\text{Enc}(\cdot)$ denotes the encoder, and $|\text{Enc}(\cdot)|$ represents stop-gradient to avoid backpropagation through the GT-masked path.

\subsubsection{Region Location Loss}
The RL loss is applied at the input feature maps of each decoder’s segmentation head to encourage coarse localization of the foreground before category-level refinement. The feature maps $F_i$ from decoder block $i$ are averaged across channels to produce attention maps, which are supervised by binarized GTs using binary cross-entropy (BCE):
\begin{equation}
L_{rl} = \sum_{i=1}^{N} BCE(mean(F_{i}), Bin(y_{i})), \label{eq:lossrl}
\end{equation}
where $N$ is the number of decoder blocks, $mean(\cdot)$ computes channel-wise average, $y_{i}$ is the supervision of the output from $i$-th decoder block, and $Bin(\cdot)$ binarizes $y_{i}$.

The overall training objective combines the segmentation loss with the FD loss:
The total loss function is formulated as:
\begin{equation}
L = L_{seg} + \alpha L_{tr} + \beta L_{rl}, \label{eq:lossoverall}
\end{equation}
where $\alpha$ and $\beta$ are set to 0.1 in all experiments.

\subsection{Progressive Pruning}
The goal of progressive pruning is to identify and remove redundant modules, enabling the model to evolve into a compact, task-specific architecture. PSP-Seg monitors the TR and RL losses at each epoch to determine pruning points. Specifically, convergence is declared if no lower loss is observed over the past 10 epochs. If the current loss improves upon the historical best, block-wise pruning is triggered. Initially, blocks are masked rather than deleted to mitigate the risk of over-pruning. If performance continues to improve, the masked blocks are permanently pruned; otherwise, they are restored, and a finer pruning step is adopted. This iterative process progressively compresses the model while preserving performance.

\subsubsection{Block-Wise Pruning}
Inspired by \cite{lu2024not}, our block-wise pruning strategy traverses each PRM to identify the least important efficient blocks. Let $f_\theta$ denote the PSP-Seg model and $PRM_l$ the $l$-th PRM. We sample a small calibration set from the training data to obtain layer input $x_l$ and output $PRM_l(x_l)$. For a given pruning step $p$, all possible combinations of $p$ blocks are evaluated. For each candidate combination, we mask the selected $p$ efficient blocks within the PRM layer, yielding a masked version denoted as $PRM_l^{(\mathcal{P})}(\cdot)$, where $\mathcal{P}$ is the set of masked efficient blocks. We recompute the output using the masked modules and measure the discrepancy from the original outputs using the Frobenius norm. The subset $\mathcal{P}^*$ with the minimal difference is selected:
\begin{equation}
\begin{split}
\min&_{\mathcal{P}} \left\| PRM_l^{(\mathcal{P})}(x_l) - PRM_l(x_l) \right\|_F \\ &\text{s.t.} \quad \mathcal{P} \subseteq \{\text{EB}_0, \ldots, \text{EB}_{n-1} \}, \ |\mathcal{P}| = p, \label{eq:pruning}
\end{split}
\end{equation}
where $EB_i$ is the $i$-th efficient block in $PRM_l$, and $n$ is the total number of blocks.
If the model’s loss stagnates or worsens post-masking (i.e., exceeds the best historical loss by 0.01), the pruned blocks are restored, and the pruning step $p$ is reduced by 1 for finer granularity. Pruning continues until $p$ reaches zero, and the same $p$ is used for all PRMs.

\renewcommand{\arraystretch}{1.1}

\begin{table*}[t]
\centering
{\small
\setlength{\tabcolsep}{0.7mm}
\begin{tabular}{cccccccc|ccccccc|ccccccc}
\hline
\multirow{2}{*}{Model} &
  \multicolumn{2}{c}{GPU M.} &
  \multicolumn{2}{c}{Time} &
  \multirow{2}{*}{\#P} &
  \multicolumn{2}{c|}{LiTS} &
  \multicolumn{2}{c}{GPU M.} &
  \multicolumn{2}{c}{Time} &
  \multirow{2}{*}{\#P} &
  \multicolumn{2}{c|}{KiTS} &
  \multicolumn{2}{c}{GPU M.} &
  \multicolumn{2}{c}{Time} &
  \multirow{2}{*}{\#P} &
  \multicolumn{2}{c}{BraTS21} \\ \cline{2-5} \cline{7-12} \cline{14-19} \cline{21-22} 
             & Tr.  & Inf. & Tr. & Inf. &       & DSC   & NSD  & Tr.  & Inf. & Tr. & Inf. &       & DSC   & NSD   & Tr.  & Inf. & Tr. & Inf. &       & DSC   & NSD   \\ \hline
nnU-Net       & 6.6  & 3.4  & 86  & 153  & 31.2  & 80.2  & 79.3 & 6.6  & 3.8  & 86  & 174  & 31.2  & 90.8  & 88.9  & 6.6  & 1.6  & 86  & 3    & 31.2  & 85.0  & 90.5  \\
CoTr         & 7.7  & 3.4  & 173 & 247  & 41.9  & 79.1  & 75.3 & 7.7  & 3.8  & 173 & 277  & 41.9  & 88.8  & 87.0  & 7.7  & 1.7  & 173 & 5    & 41.9  & 83.4  & 89.5  \\
nnFormer     & 7.4  & 4.1  & 65  & 128  & 151.0 & 80.3  & 78.7 & 7.4  & 4.6  & 65  & 136  & 151.0 & 88.6  & 86.9  & 7.4  & 2.4  & 65  & 3    & 151.0 & 84.4  & 90.3  \\
SwinUNETR-V2 & 21.4 & 4.0  & 147 & 240  & 72.8  & 78.1  & 75.8 & 21.4 & 4.4  & 147 & 271  & 72.8  & 89.2  & 87.5  & 21.4 & 1.8  & 147 & 5    & 72.8  & 83.8  & 89.4  \\
U-Mamba      & 11.8 & 3.9  & 130 & 233  & 70.0  & 80.4  & 78.8 & 11.8 & 4.3  & 130 & 275  & 70.0  & 90.5  & 88.4  & 11.8 & 1.8  & 130 & 5    & 70.0  & 84.7  & 90.0  \\
STU-Net$\dagger$     & 8.0  & 3.7  & 88  & 172  & 58.2  & 80.9  & 79.5 & 8.0  & 4.1  & 88  & 187  & 58.2  & 91.0  & 88.3  & 8.0  & 1.9  & 88  & 4    & 58.2  & 84.9  & 90.4  \\
UniMiSS+$\dagger$    & 6.3  & 3.1  & 61  & 102  & 54.2  & 80.8  & 80.1 & 6.3  & 3.5  & 61  & 126  & 54.2  & 90.4  & 88.8  & 6.3  & 1.5  & 61  & 3    & 54.2  & 84.2  & 89.9  \\ \hline
DMF-Net      & 3.6  & 2.5  & 36  & 57   & 3.9   & 78.2  & 76.0 & 3.6  & 3.0  & 36  & 59   & 3.9   & 83.9  & 81.4  & 3.6  & 0.8  & 36  & 1    & 3.9   & 83.8  & 89.8  \\
SegFormer3D  & 2.1  & 2.2  & 26  & 36   & 4.5   & 78.3  & 76.2 & 2.1  & 2.7  & 26  & 34   & 4.5   & 85.9  & 83.6  & 2.1  & 0.6  & 26  & 1    & 4.5   & 80.4  & 88.6  \\
LHU-Net      & 5.3  & 2.8  & 232 & 200  & 13.0  & 77.8  & 75.9 & 5.3  & 3.2  & 232 & 221  & 13.0  & 87.3  & 85.3  & 5.3  & 1.0  & 232 & 4    & 13.0  & 84.0  & 89.6  \\
Slim   UNETR & 1.2  & 2.1  & 34  & 86   & 1.8   & 55.0  & 40.2 & 1.2  & 2.5  & 34  & 86   & 1.8   & 76.4  & 70.7  & 1.2  & 0.5  & 34  & 2    & 1.8   & 71.8  & 81.0  \\
EffiDec3D    & 3.7  & 2.5  & 126 & 129  & 3.2   & 78.4  & 76.1 & 3.7  & 2.9  & 126 & 172  & 3.2   & 87.3  & 84.3  & 3.7  & 0.8  & 126 & 3    & 3.2   & 79.8  & 89.4  \\
MLRU++       & 3.8  & 2.8  & 61  & 120  & 46.1  & 79.0  & 76.1 & 3.8  & 3.3  & 61  & 133  & 46.1  & 83.8  & 80.9  & 3.8  & 1.0  & 61  & 3    & 46.1  & 84.2  & 90.0  \\
PSP-Seg-S$\dagger$     & 3.8  & 2.5  & 52  & 60   & 4.3   & 79.2  & 76.5 & 3.7  & 2.9  & 47  & 74   & 4.2   & 88.7  & 85.5  & 3.7  & 0.8  & 61  & 2    & 5.1   & 84.3* & 89.9  \\
PSP-Seg-B$\dagger$     & 5.0  & 2.7  & 62  & 111  & 10.3  & 80.3* & 77.4 & 5.5  & 3.1  & 74  & 171  & 10.7  & 89.3  & 87.5  & 5.2  & 0.9  & 68  & 3    & 11.7  & 84.5* & 90.1* \\
PSP-Seg-L$\dagger$     & 5.2  & 2.6  & 63  & 148  & 19.8  & 81.0* & 78.5 & 6.3  & 3.1  & 90  & 180  & 23.4  & 90.6* & 88.2* & 5.3  & 1.0  & 70  & 3    & 24.2  & 85.2* & 90.8* \\ \hline
\end{tabular}
}
\caption{Resource consumption and segmentation performance on the LiTS, KiTS, and BraTS21 datasets. 'Tr.' and 'Inf.' denote training and inference, respectively. 'GPU M.' refers to GPU memory usage (GB, $\downarrow$), and 'Time' indicates the average training time per epoch or inference time per image (seconds, $\downarrow$). '\#P' denotes the number of model parameters (millions, $\downarrow$). The average Dice Similarity Coefficient (DSC, \% $\uparrow$) and average Normalized Surface Dice (NSD, \% $\uparrow$) are reported. $\dagger$ indicates models with pre-trained weights. The Wilcoxon signed-rank test is employed on our PSP-Seg results and the best result for each metric. Results without statistical significance ($p>0.05$) are marked with *.}
\label{tab:lits&kits&brats}
\end{table*}

\section{Experiments and Results}
\subsection{Implementations}
PSP-Seg is implemented based on the nnU-Net V2 framework \cite{isensee2021nnu}.  Following STU-Net \cite{huang2023stu}, we pre-train the base model on the TotalSegmentator v2 dataset  \cite{wasserthal2023totalsegmentator} for 1,000 epochs with a batch size of 2 in a supervised manner.  The optimizer is set to SGD with an initial learning rate of 0.001 to ensure robust training.
For target task learning, we use the AdamW optimizer with a batch size of 2 and an initial learning rate of 0.0001. The patch size is set to $128 \times 128 \times 128$, and the total number of training epochs is 1000 with a total of 250,000 iterations. During each pruning step, we randomly sample 100 images from the training set to evaluate pruning impact. The initial pruning step size $p$ is set to 2 for PSP-Seg-L and 1 for PSP-Seg-B and PSP-Seg-S.
To accommodate varying resource budgets, we configure three model variants by adjusting the model depth (denoted as $d$) and the number of efficient blocks ($m$) per PRM: PSP-Seg-S (d=5, m=4), PSP-Seg-B (d=5, m=7), and PSP-Seg-L (d=6, m=7). More detail of our base model can found in Appendix \ref{app:implementation}.

\subsection{Evaluation Metrics}
The segmentation performance is evaluated using the Dice Similarity Coefficient (DSC), which measures the overlap between predicted and ground truth regions, and the Normalized Surface Dice (NSD) with a 2mm tolerance to evaluate boundary alignment.  GPU memory cost is measured by directly accessing GPU usage. For time consumption, we record the average training time per epoch and the average inference time per image. Additionally, we report the final number of model parameters after pruning for PSP-Seg.

\subsection{Datasets}
In this study, we evaluated PSP-Seg on five public datasets: LiTS \cite{bilic2023lits}, KiTS \cite{heller2021kits}, BraTS21 \cite{baid2021brats21}, Hepatic Vessel \cite{antonelli2022medical}, and Colon \cite{antonelli2022medical}. 
All datasets are preprocessed using the nnU-Net pipeline and randomly split into training and test sets with an 80:20 ratio. Additionally, the TotalSegmentator v2 dataset \cite{wasserthal2023totalsegmentator} is used exclusively for supervised pre-training.
Further dataset details and statistics are provided in Appendix \ref{app:datasets}.

% The LiTS \cite{bilic2023lits} dataset comprises 131 contrast-enhanced 3D abdominal CT scans with liver and tumor annotations. The KiTs \cite{heller2021kits} includes CT scans of kidney cancer patients who underwent nephrectomy, annotated with kidneys and kidney tumors. The BraTS21 dataset \cite{baid2021brats21} annotates brain tumors across four MRI modalities (T1, T1-weighted, T2-weighted, and T2-FLAIR), providing segmentation for all 1,251 paired multi-modal images on peritumoral edematous/invaded tissue, the necrotic tumor core, and the Gd-enhancing tumor. The Hepatic Vessel and Colon datasets are from Medical Segmentation Decathlon(MSD) \cite{antonelli2022medical}. The Hepatic Vessel dataset consists of 303 liver CT scans with the task of segmenting hepatic vessels and tumors. The Colon dataset includes 126 contrast-enhanced CT images focused on segmenting primary colon cancer tumors. 

\begin{table*}[t]
\centering
{\small
\setlength{\tabcolsep}{1.5mm}
\begin{tabular}{cccccccc|ccccccc}
\hline
\multirow{2}{*}{Model} &
  \multicolumn{2}{c}{GPU M.} &
  \multicolumn{2}{c}{Time} &
  \multirow{2}{*}{\#P} &
  \multicolumn{2}{c|}{Colon} &
  \multicolumn{2}{c}{GPU M.} &
  \multicolumn{2}{c}{Time} &
  \multirow{2}{*}{\#P} &
  \multicolumn{2}{c}{HepaticVessel} \\ \cline{2-5} \cline{7-12} \cline{14-15} 
             & Tr.  & Inf. & Tr. & Inf. &       & DSC   & NSD   & Tr.  & Inf. & Tr. & Inf. &       & DSC   & NSD   \\ \hline
nnU-Net       & 6.6  & 2.1  & 86  & 50   & 31.2  & 37.9  & 37.7  & 6.6  & 2.2  & 86  & 47   & 31.2  & 69.1  & 72.9  \\
CoTr         & 7.7  & 2.1  & 173 & 67   & 41.9  & 32.4  & 32.9  & 7.7  & 2.2  & 173 & 66   & 41.9  & 61.9  & 62.1  \\
nnFormer     & 7.4  & 2.8  & 65  & 51   & 151.0 & 32.7  & 33.5  & 7.4  & 2.9  & 65  & 48   & 151.0 & 67.2  & 70.2  \\
SwinUNETR-V2 & 21.4 & 2.6  & 147 & 73   & 72.8  & 32.8  & 32.2  & 21.4 & 2.8  & 147 & 71   & 72.8  & 66.8  & 68.2  \\
U-Mamba      & 11.8 & 2.5  & 130 & 78   & 70.0  & 35.9  & 35.6  & 11.8 & 2.7  & 130 & 70   & 70.0  & 69.7  & 72.1  \\
STU-Net$\dagger$     & 8.0  & 2.4  & 88  & 72   & 58.2  & 40.8  & 41.0  & 8.0  & 2.5  & 88  & 55   & 58.2  & 69.5  & 72.5  \\
UniMiSS+$\dagger$    & 6.3  & 1.7  & 61  & 30   & 54.2  & 39.5  & 40.2  & 6.3  & 1.9  & 61  & 30   & 54.2  & 69.6  & 72.3  \\ \hline
DMF-Net      & 3.6  & 1.2  & 36  & 33   & 3.9   & 31.6  & 30.7  & 3.6  & 1.4  & 36  & 19   & 3.9   & 64.6  & 67.4  \\
SegFormer3D  & 2.1  & 1.0  & 26  & 21   & 4.5   & 29.7  & 28.4  & 2.1  & 1.1  & 26  & 19   & 4.5   & 64.7  & 67.5  \\
LHU-Net      & 5.3  & 1.5  & 232 & 71   & 13.0  & 36.2  & 36.7  & 5.3  & 1.7  & 232 & 72   & 13.0  & 65.9  & 68.9  \\
SlimUNETR    & 1.2  & 0.8  & 34  & 28   & 1.8   & 20.3  & 18.8  & 1.2  & 1.0  & 34  & 22   & 1.8   & 39.2  & 38.3  \\
EffiDec3D    & 3.7  & 1.2  & 126 & 39   & 3.2   & 28.2  & 27.3  & 3.7  & 1.2  & 126 & 37   & 3.2   & 65.6  & 69.1  \\
MLRU++       & 3.8  & 1.5  & 61  & 55   & 46.1  & 29.2  & 29.4  & 3.8  & 1.7  & 61  & 45   & 46.1  & 67.0  & 69.8  \\
PSP-Seg-S$\dagger$     & 3.6  & 1.2  & 58  & 34   & 5.4   & 41.3* & 41.3* & 3.6  & 1.4  & 45  & 20   & 5.2   & 69.0* & 71.7  \\
PSP-Seg-B$\dagger$     & 4.9  & 1.3  & 61  & 35   & 9.0   & 40.7* & 41.3* & 5.0  & 1.5  & 64  & 35   & 9.6   & 68.3  & 71.4  \\
PSP-Seg-L$\dagger$     & 5.2  & 1.4  & 62  & 39   & 19.8  & 41.2* & 42.3* & 5.2  & 1.6  & 68  & 38   & 22.0  & 69.8* & 72.5* \\ \hline
\end{tabular}
}
\caption{Resource consumption and segmentation performance on the Hepatic Vessel and Colon datasets.
All evaluation settings and metrics follow those described in Table \ref{tab:lits&kits&brats}.}
\label{tab:hv&colon}
\end{table*}

\subsection{Comparing to Advanced and Efficient Segmentation Models}
We compared our proposed PSP-Seg with seven state-of-the-art segmentation models, encompassing CNN-based models such as nnU-Net \cite{isensee2021nnu} and STU-Net \cite{huang2023stu}, Transformer-based models like nnFormer \cite{zhou2021nnformer}, SwinUNETR-V2 \cite{he2023swinunetr}, and UniMiSS+ \cite{xie2024unimiss+}, the Mamba-based method U-Mamba \cite{ma2024umamba}, and hybrid approaches such as CoTr \cite{xie2021cotr}, and six recent efficient models, including DMF-Net \cite{chen2019dmfnet}, SegFormer3D \cite{perera2024segformer3d}, LHU-Net \cite{sadegheih2024lhunet}, Slim UNETR \cite{pang2024slimunetr}, EffiDec3D \cite{rahman2025effidec3d}, MLRU++ \cite{yadav2025mlru++}. For clarity, we used nnU-Net V2 and the STU-Net-B variant in our comparisons. U-Mamba refers specifically to the U-Mamba\_Bot model. UniMiSS+ and STU-Net were fine-tuned using pre-trained weights, while all other models were trained from scratch.
To ensure fair comparisons, all models were implemented within the nnU-Net V2 framework \cite{isensee2021nnu}.
Experimental results, summarized in Tables \ref{tab:lits&kits&brats} and \ref{tab:hv&colon}, demonstrate the following:
(1) Among advanced segmentation models, STU-Net achieved the highest Dice scores on four out of five datasets. On the Hepatic Vessel dataset, however, U-Mamba and UniMiSS+ slightly outperformed it.
(2) Among the efficient segmentation models, our PSP-Seg variants (PSP-Seg-S, PSP-Seg-B, and PSP-Seg-L) consistently delivered the best trade-off between segmentation accuracy and computational efficiency. PSP-Seg models achieved top performance across all datasets while maintaining low resource usage. For example, compared to MLRU++, PSP-Seg-S achieved comparable GPU memory consumption, required less inference time, and used fewer parameters. Despite these lower resource requirements, PSP-Seg-S outperformed MLRU++ by 0.2\%, 4.9\%, 0.1\%, 12.1\%, and 2.0\% in Dice scores across the evaluated datasets. While Slim UNETR offered the lowest resource usage across all metrics, it failed to maintain competitive segmentation performance.
(3) Overall, the PSP-Seg series delivered segmentation performance that was statistically comparable to the best-performing models ($p > 0.05$), while significantly reducing GPU memory consumption, inference time, and parameter number. These results underscore the effectiveness and efficiency of our proposed PSP-Seg framework. Detailed experimental results are provided in Appendix \ref{app:Main exp}.

\subsection{Ablation Studies}
We conducted ablation studies on each component of PSP-Seg, including the Parallel Redundant Module (PRM), Functional Decoupling Loss (FD Loss), Progressive Pruning (PSP), and pre-training, using the LiTS dataset. PSP-Seg-L was used as the baseline model, with additional results on other model variants provided in Appendix \ref{app:ablation}. As summarized in Table \ref{tab:ablation}, the findings are as follows:
(1) The baseline model, which consists solely of $1 \times 1 \times 1$ efficient blocks without any of our proposed components, requires minimal computational resources but yields suboptimal segmentation performance.
(2) Introducing the PRM significantly improves segmentation accuracy, albeit at the cost of increased resource consumption due to redundant computations.
(3) By incorporating FD Loss and PSP, the model is able to selectively prune redundant modules, reducing resource usage while maintaining performance.
(4) Finally, with pre-training, individual modules become more expressive, allowing PSP to prune even more aggressively. This leads to further reductions in computational cost and, notably, improved segmentation accuracy.

\begin{table}[t]
\centering
{\small
\setlength{\tabcolsep}{0.7mm}
\begin{tabular}{ccccccccccc}
\hline
\multirow{2}{*}{PRM} &
  \multirow{2}{*}{FD Loss} &
  \multirow{2}{*}{PSP} &
  \multirow{2}{*}{PT} &
  \multicolumn{2}{c}{GPU Mem.} &
  \multicolumn{2}{c}{Time} &
  \multirow{2}{*}{\#P} &
  \multicolumn{2}{c}{AVG} \\ \cline{5-8} \cline{10-11} 
  &   &   &   & Tr. & Inf. & Tr. & Inf. &      & DSC           & NSD           \\ \hline
\ding{55} & \ding{55} & \ding{55} & \ding{55} & \textbf{4.6} & \textbf{2.6}  & \textbf{42}  & \textbf{113}  & \textbf{9.9}  & 78.1          & 75.3          \\
\ding{51} & \ding{55} & \ding{55} & \ding{55} & 7.8 & 2.8  & 100 & 258  & 33.1 & 80.1          & 77.7          \\
\ding{51} & \ding{51} & \ding{55} & \ding{55} & 7.8 & 2.8  & 101 & 260  & 33.1 & 80.1          & 77.3          \\
\ding{51} & \ding{51} & \ding{51} & \ding{55} & 6.4 & 2.7  & 89  & 163  & 29.4 & 80.4          & 77.5          \\
\ding{51} & \ding{51} & \ding{51} & \ding{51} & 5.2 & \textbf{2.6}  & \textbf{63}  & 148  & 19.8 & \textbf{81.0} & \textbf{78.5} \\ \hline
\end{tabular}
}
\caption{Ablation study of the PRM, FD Loss, PSP, and pre-training components in PSP-Seg on the LiTS dataset. PRM refers to the Parallel Redundant Module, FD Loss denotes the Functional Decoupling Loss, PSP stands for Progressive Pruning, and PT indicates the use of pre-trained weights. The best performance for each metric is highlighted in \textbf{bold}.}
\label{tab:ablation}
\end{table}

\begin{table}[t]
\centering
{\small
\setlength{\tabcolsep}{0.8mm}
\begin{tabular}{cccccccc}
\hline
\multirow{2}{*}{Pre-train} & \multirow{2}{*}{Method} & \multicolumn{2}{c}{Liver} & \multicolumn{2}{c}{Tumor} & \multicolumn{2}{c}{AVG}       \\ \cline{3-8}
                    &         & DSC           & NSD           & DSC           & NSD           & DSC           & NSD           \\ \hline
\multirow{2}{*}{\ding{55}} & Re-train & 94.9          & 86.0          & 63.6          & 66.8          & 79.2          & 76.4          \\
                    & PSP & \textbf{95.6} & \textbf{86.2} & \textbf{65.1} & \textbf{68.5} & \textbf{80.4} & \textbf{77.5} \\ \hline
\multirow{2}{*}{\ding{51}} & Re-train & 95.3          & 86.4          & \textbf{66.1} & 69.1          & 80.7          & 77.8          \\
                    & PSP & \textbf{96.1} & \textbf{87.3} & 65.9          & \textbf{69.6} & \textbf{81.0} & \textbf{78.5} \\ \hline
\end{tabular}
}
\caption{Segmentation performance of the re-trained model and the model generated by PSP. PSP-Seg-L serves as the baseline. The best performance is highlighted in \textbf{bold}.}
\label{tab:retrain}
\end{table}

\begin{figure}[t]
\centerline{\includegraphics[width=0.95\linewidth]{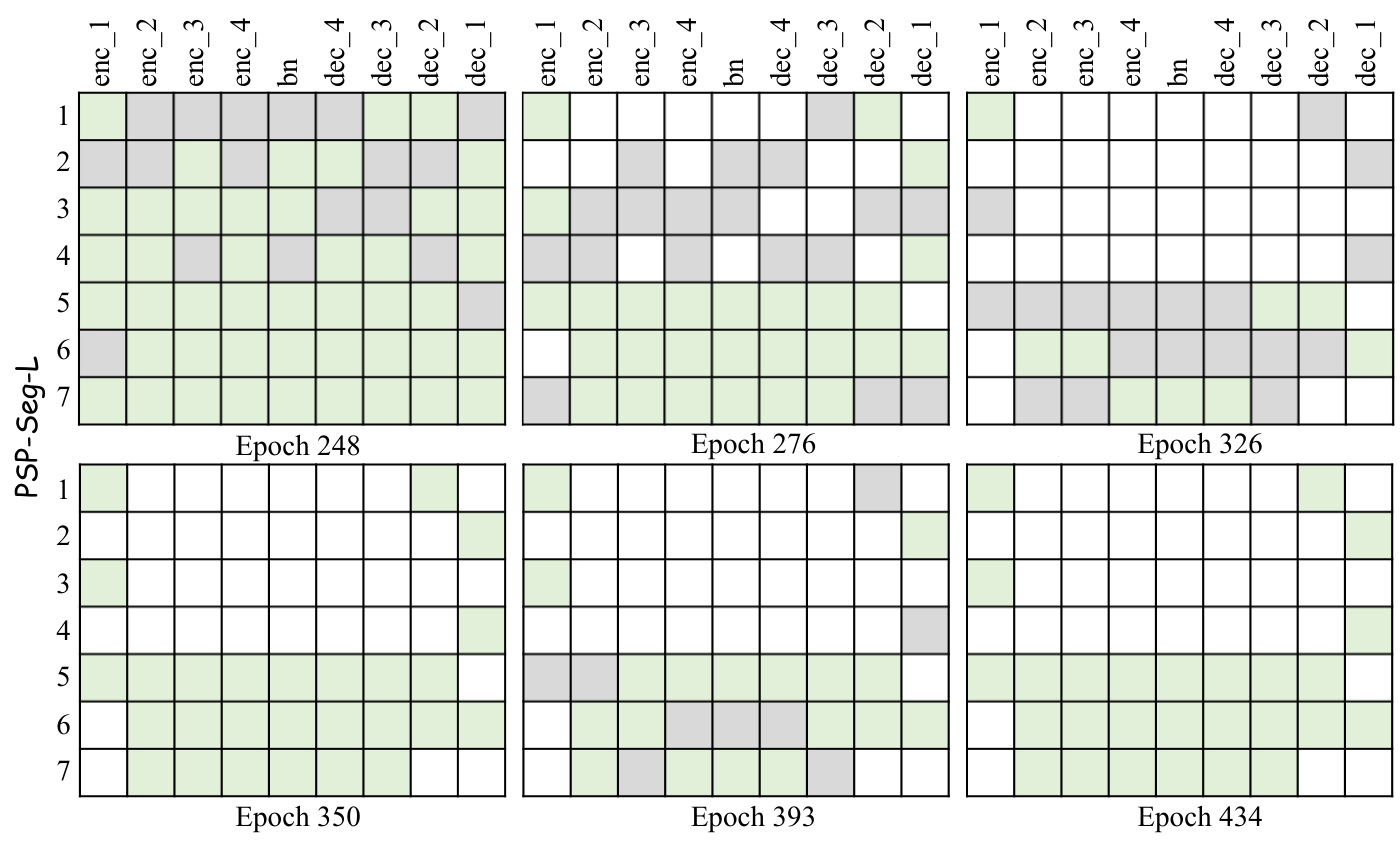}}
\caption{Visualization of PSP-Seg-L across six training epochs on the LiTS dataset. Green, gray, and white patches indicate retained, masked, and pruned efficient blocks, respectively. The labels `1–7' denote efficient blocks with kernel sizes of $1\times1\times1$, $1\times1\times3$, $1\times3\times1$, $3\times1\times1$, $1\times3\times3$, $3\times1\times3$, and $3\times3\times1$, respectively. The notations $enc_i$, $bn$, and $dec_i$ refer to the $i$-th encoder block, the bottleneck layer, and the $i$-th decoder block, respectively.}
\label{fig:prune_progress}
\end{figure}

\section{Discussions}
\subsection{Re-training Pruned Model}
PSP-Seg progressively prunes a redundant base model to obtain a compact version. In this section, we further investigate whether the compact model alone is well-suited for the segmentation task. To this end, we re-trained the compact model on the Liver dataset, both with and without pre-trained weights. The baseline model is denoted as PSP-Seg-L. The corresponding results are presented in Table \ref{tab:retrain}, and additional experiments on different model variants are provided in Appendix \ref{app:retrain}.
We draw two key observations: (1) Pre-training once again proves to be beneficial for performance. (2) Compact models re-trained from scratch consistently underperform compared to the counterparts obtained through PSP pruning process, across all model sizes. This performance gap may stem from the enhanced representation capacity of the original redundant model. PSP-Seg gradually encourages the retained modules to inherit knowledge from the pruned ones, resulting in more powerful components than those learned directly in the compact model.

\subsection{Which Modules Should Be Pruned?}
We visualize the model structure throughout the progressive pruning process on the LiTS dataset. As illustrated in Fig. \ref{fig:prune_progress}, PSP-Seg-L serves as the baseline model. In the figure, green, gray, and white patches represent retained, masked, and pruned efficient blocks, respectively. It can be observed that blocks with small kernel sizes tend to be preserved at both the early and late stages of the forward pass, while blocks with larger kernel sizes dominate the middle stages. Additional visualizations are provided in Appendix \ref{app:pruneprogress}. This pattern aligns with the common intuition: low-level (local) features are primarily captured in the shallow encoder and deep decoder layers, whereas high-level (semantic) information is extracted in the central parts of the model.

\section{Conclusion}
In this paper, we propose a progressive pruning framework, termed PSP-Seg, for efficient 3D medical image segmentation. PSP-Seg is designed to identify and eliminate redundant network components during training, ultimately yielding a compact, task-specific model. The process begins with a redundant model, which is iteratively pruned using a progressive strategy based on functional decoupling loss and block-wise pruning. We evaluate PSP-Seg on five challenging 3D medical segmentation benchmarks, assessing both segmentation performance and computational efficiency. Extensive experiments demonstrate that PSP-Seg achieves a favorable trade-off between performance and resource consumption. In future work, we plan to investigate the scaling laws governing PSP-Seg and explore its integration with more advanced backbone architectures.

\section{Acknowledgment}
This work was supported in part by the National Natural Science Foundation of China under Grant 92470101 and Grant 62171377, in part by the "Pioneer" and "Leading Goose" R\&D Program of Zhejiang, China, under Grant 2025C01201(SD2), in part by the Ningbo Clinical Research Center for Medical Imaging under Grant 2021L003 (Open Project 2022LYKFZD06), and in part by the Innovation Foundation for Doctor Dissertation of Northwestern Polytechnical University under Grants CX2024016.

\bibliography{aaai2026}

\begin{thebibliography}{39}
\providecommand{\natexlab}[1]{#1}

\bibitem[{Antonelli et~al.(2022)Antonelli, Reinke, Bakas, Farahani, Kopp-Schneider, Landman, Litjens, Menze, Ronneberger, Summers et~al.}]{antonelli2022medical}
Antonelli, M.; Reinke, A.; Bakas, S.; Farahani, K.; Kopp-Schneider, A.; Landman, B.~A.; Litjens, G.; Menze, B.; Ronneberger, O.; Summers, R.~M.; et~al. 2022.
\newblock The medical segmentation decathlon.
\newblock \emph{Nature communications}, 13(1): 4128.

\bibitem[{Azad et~al.(2024)Azad, Aghdam, Rauland, Jia, Avval, Bozorgpour, Karimijafarbigloo, Cohen, Adeli, and Merhof}]{azad2024medical}
Azad, R.; Aghdam, E.~K.; Rauland, A.; Jia, Y.; Avval, A.~H.; Bozorgpour, A.; Karimijafarbigloo, S.; Cohen, J.~P.; Adeli, E.; and Merhof, D. 2024.
\newblock Medical image segmentation review: The success of u-net.
\newblock \emph{IEEE Transactions on Pattern Analysis and Machine Intelligence}.

\bibitem[{Baid et~al.(2021)Baid, Ghodasara, Mohan, Bilello, Calabrese, Colak, Farahani, Kalpathy-Cramer, Kitamura, Pati et~al.}]{baid2021brats21}
Baid, U.; Ghodasara, S.; Mohan, S.; Bilello, M.; Calabrese, E.; Colak, E.; Farahani, K.; Kalpathy-Cramer, J.; Kitamura, F.~C.; Pati, S.; et~al. 2021.
\newblock The rsna-asnr-miccai brats 2021 benchmark on brain tumor segmentation and radiogenomic classification.
\newblock \emph{arXiv preprint arXiv:2107.02314}.

\bibitem[{Bilic et~al.(2023)Bilic, Christ, Li, Vorontsov, Ben-Cohen, Kaissis, Szeskin, Jacobs, Mamani, Chartrand et~al.}]{bilic2023lits}
Bilic, P.; Christ, P.; Li, H.~B.; Vorontsov, E.; Ben-Cohen, A.; Kaissis, G.; Szeskin, A.; Jacobs, C.; Mamani, G. E.~H.; Chartrand, G.; et~al. 2023.
\newblock The liver tumor segmentation benchmark (lits).
\newblock \emph{Medical image analysis}, 84: 102680.

\bibitem[{Chen et~al.(2019)Chen, Liu, Ding, Zheng, and Li}]{chen2019dmfnet}
Chen, C.; Liu, X.; Ding, M.; Zheng, J.; and Li, J. 2019.
\newblock 3D dilated multi-fiber network for real-time brain tumor segmentation in MRI.
\newblock In \emph{Medical Image Computing and Computer Assisted Intervention--MICCAI 2019: 22nd International Conference, Shenzhen, China, October 13--17, 2019, Proceedings, Part III 22}, 184--192. Springer.

\bibitem[{Chen et~al.(2021)Chen, Lu, Yu, Luo, Adeli, Wang, Lu, Yuille, and Zhou}]{chen2021transunet}
Chen, J.; Lu, Y.; Yu, Q.; Luo, X.; Adeli, E.; Wang, Y.; Lu, L.; Yuille, A.~L.; and Zhou, Y. 2021.
\newblock Transunet: Transformers make strong encoders for medical image segmentation.
\newblock \emph{arXiv preprint arXiv:2102.04306}.

\bibitem[{Chen et~al.(2022)Chen, Huang, Xie, Jiao, Jiang, Zhou, Li, and Wei}]{chen2022task}
Chen, T.; Huang, S.; Xie, Y.; Jiao, B.; Jiang, D.; Zhou, H.; Li, J.; and Wei, F. 2022.
\newblock Task-specific expert pruning for sparse mixture-of-experts.
\newblock \emph{arXiv preprint arXiv:2206.00277}.

\bibitem[{Cheng, Zhang, and Shi(2024)}]{cheng2024prunesurvey}
Cheng, H.; Zhang, M.; and Shi, J.~Q. 2024.
\newblock A survey on deep neural network pruning: Taxonomy, comparison, analysis, and recommendations.
\newblock \emph{IEEE Transactions on Pattern Analysis and Machine Intelligence}.

\bibitem[{Cho, Adya, and Naik(2023)}]{cho2023pdp}
Cho, M.; Adya, S.; and Naik, D. 2023.
\newblock PDP: Parameter-free differentiable pruning is all you need.
\newblock \emph{Advances in Neural Information Processing Systems}, 36: 45833--45855.

\bibitem[{Frantar and Alistarh(2023)}]{frantar2023sparsegpt}
Frantar, E.; and Alistarh, D. 2023.
\newblock Sparsegpt: Massive language models can be accurately pruned in one-shot.
\newblock In \emph{International Conference on Machine Learning}, 10323--10337. PMLR.

\bibitem[{Gao et~al.(2022)Gao, Zhou, Liu, Yan, Zhang, and Metaxas}]{gao2022medformer}
Gao, Y.; Zhou, M.; Liu, D.; Yan, Z.; Zhang, S.; and Metaxas, D.~N. 2022.
\newblock A data-scalable transformer for medical image segmentation: architecture, model efficiency, and benchmark.
\newblock \emph{arXiv preprint arXiv:2203.00131}.

\bibitem[{Hatamizadeh et~al.(2021)Hatamizadeh, Nath, Tang, Yang, Roth, and Xu}]{hatamizadeh2021swin}
Hatamizadeh, A.; Nath, V.; Tang, Y.; Yang, D.; Roth, H.~R.; and Xu, D. 2021.
\newblock Swin unetr: Swin transformers for semantic segmentation of brain tumors in mri images.
\newblock In \emph{International MICCAI brainlesion workshop}, 272--284. Springer.

\bibitem[{He et~al.(2019)He, Liu, Wang, Hu, and Yang}]{he2019fpgm}
He, Y.; Liu, P.; Wang, Z.; Hu, Z.; and Yang, Y. 2019.
\newblock Filter pruning via geometric median for deep convolutional neural networks acceleration.
\newblock In \emph{Proceedings of the IEEE/CVF conference on computer vision and pattern recognition}, 4340--4349.

\bibitem[{He et~al.(2023)He, Nath, Yang, Tang, Myronenko, and Xu}]{he2023swinunetr}
He, Y.; Nath, V.; Yang, D.; Tang, Y.; Myronenko, A.; and Xu, D. 2023.
\newblock Swinunetr-v2: Stronger swin transformers with stagewise convolutions for 3d medical image segmentation.
\newblock In \emph{International Conference on Medical Image Computing and Computer-Assisted Intervention}, 416--426. Springer.

\bibitem[{Heller et~al.(2021)Heller, Isensee, Maier-Hein, Hou, Xie, Li, Nan, Mu, Lin, Han et~al.}]{heller2021kits}
Heller, N.; Isensee, F.; Maier-Hein, K.~H.; Hou, X.; Xie, C.; Li, F.; Nan, Y.; Mu, G.; Lin, Z.; Han, M.; et~al. 2021.
\newblock The state of the art in kidney and kidney tumor segmentation in contrast-enhanced CT imaging: Results of the KiTS19 challenge.
\newblock \emph{Medical image analysis}, 67: 101821.

\bibitem[{Huang et~al.(2023)Huang, Wang, Deng, Ye, Su, Sun, He, Gu, Gu, Zhang et~al.}]{huang2023stu}
Huang, Z.; Wang, H.; Deng, Z.; Ye, J.; Su, Y.; Sun, H.; He, J.; Gu, Y.; Gu, L.; Zhang, S.; et~al. 2023.
\newblock Stu-net: Scalable and transferable medical image segmentation models empowered by large-scale supervised pre-training.
\newblock \emph{arXiv preprint arXiv:2304.06716}.

\bibitem[{Isensee et~al.(2021)Isensee, Jaeger, Kohl, Petersen, and Maier-Hein}]{isensee2021nnu}
Isensee, F.; Jaeger, P.~F.; Kohl, S.~A.; Petersen, J.; and Maier-Hein, K.~H. 2021.
\newblock nnU-Net: a self-configuring method for deep learning-based biomedical image segmentation.
\newblock \emph{Nature methods}, 18(2): 203--211.

\bibitem[{Li et~al.(2019)Li, Wu, Yang, Fan, Zhang, and Liu}]{li2019fcf}
Li, T.; Wu, B.; Yang, Y.; Fan, Y.; Zhang, Y.; and Liu, W. 2019.
\newblock Compressing convolutional neural networks via factorized convolutional filters.
\newblock In \emph{Proceedings of the IEEE/CVF conference on computer vision and pattern recognition}, 3977--3986.

\bibitem[{Liu et~al.(2021)Liu, Chen, Atashgahi, Chen, Sokar, Mocanu, Pechenizkiy, Wang, and Mocanu}]{liu2021freetickets}
Liu, S.; Chen, T.; Atashgahi, Z.; Chen, X.; Sokar, G.; Mocanu, E.; Pechenizkiy, M.; Wang, Z.; and Mocanu, D.~C. 2021.
\newblock Deep ensembling with no overhead for either training or testing: The all-round blessings of dynamic sparsity.
\newblock \emph{arXiv preprint arXiv:2106.14568}.

\bibitem[{Lu et~al.(2024)Lu, Liu, Xu, Zhou, Huang, Zhang, Yan, and Li}]{lu2024not}
Lu, X.; Liu, Q.; Xu, Y.; Zhou, A.; Huang, S.; Zhang, B.; Yan, J.; and Li, H. 2024.
\newblock Not all experts are equal: Efficient expert pruning and skipping for mixture-of-experts large language models.
\newblock \emph{arXiv preprint arXiv:2402.14800}.

\bibitem[{Ma, Li, and Wang(2024)}]{ma2024umamba}
Ma, J.; Li, F.; and Wang, B. 2024.
\newblock U-mamba: Enhancing long-range dependency for biomedical image segmentation.
\newblock \emph{arXiv preprint arXiv:2401.04722}.

\bibitem[{Ma et~al.(2018)Ma, Zhang, Zheng, and Sun}]{ma2018shufflenetv2}
Ma, N.; Zhang, X.; Zheng, H.-T.; and Sun, J. 2018.
\newblock Shufflenet v2: Practical guidelines for efficient cnn architecture design.
\newblock In \emph{Proceedings of the European conference on computer vision (ECCV)}, 116--131.

\bibitem[{Maas et~al.(2013)Maas, Hannun, Ng et~al.}]{maas2013leakyrelu}
Maas, A.~L.; Hannun, A.~Y.; Ng, A.~Y.; et~al. 2013.
\newblock Rectifier nonlinearities improve neural network acoustic models.
\newblock In \emph{Proc. icml}, volume~30, 3. Atlanta, GA.

\bibitem[{Men et~al.(2024)Men, Xu, Zhang, Wang, Lin, Lu, Han, and Chen}]{men2024shortgpt}
Men, X.; Xu, M.; Zhang, Q.; Wang, B.; Lin, H.; Lu, Y.; Han, X.; and Chen, W. 2024.
\newblock Shortgpt: Layers in large language models are more redundant than you expect.
\newblock \emph{arXiv preprint arXiv:2403.03853}.

\bibitem[{Pang et~al.(2024)Pang, Liang, Huang, Chen, Li, Li, Huang, and Wang}]{pang2024slimunetr}
Pang, Y.; Liang, J.; Huang, T.; Chen, H.; Li, Y.; Li, D.; Huang, L.; and Wang, Q. 2024.
\newblock Slim UNETR: Scale Hybrid Transformers to Efficient 3D Medical Image Segmentation Under Limited Computational Resources.
\newblock \emph{IEEE Transactions on Medical Imaging}, 43(3): 994--1005.

\bibitem[{Perera, Navard, and Yilmaz(2024)}]{perera2024segformer3d}
Perera, S.; Navard, P.; and Yilmaz, A. 2024.
\newblock Segformer3d: an efficient transformer for 3d medical image segmentation.
\newblock In \emph{Proceedings of the IEEE/CVF Conference on Computer Vision and Pattern Recognition}, 4981--4988.

\bibitem[{Rahman and Marculescu(2025)}]{rahman2025effidec3d}
Rahman, M.~M.; and Marculescu, R. 2025.
\newblock EffiDec3D: An Optimized Decoder for High-Performance and Efficient 3D Medical Image Segmentation.
\newblock In \emph{Proceedings of the Computer Vision and Pattern Recognition Conference}, 10435--10444.

\bibitem[{Rayed et~al.(2024)Rayed, Islam, Niha, Jim, Kabir, and Mridha}]{rayed2024deep}
Rayed, M.~E.; Islam, S.~S.; Niha, S.~I.; Jim, J.~R.; Kabir, M.~M.; and Mridha, M. 2024.
\newblock Deep learning for medical image segmentation: State-of-the-art advancements and challenges.
\newblock \emph{Informatics in Medicine Unlocked}, 101504.

\bibitem[{Ronneberger, Fischer, and Brox(2015)}]{ronneberger2015unet}
Ronneberger, O.; Fischer, P.; and Brox, T. 2015.
\newblock U-net: Convolutional networks for biomedical image segmentation.
\newblock In \emph{Medical image computing and computer-assisted intervention--MICCAI 2015: 18th international conference, Munich, Germany, October 5-9, 2015, proceedings, part III 18}, 234--241. Springer.

\bibitem[{Roy et~al.(2023)Roy, Koehler, Ulrich, Baumgartner, Petersen, Isensee, Jaeger, and Maier-Hein}]{roy2023mednext}
Roy, S.; Koehler, G.; Ulrich, C.; Baumgartner, M.; Petersen, J.; Isensee, F.; Jaeger, P.~F.; and Maier-Hein, K.~H. 2023.
\newblock Mednext: transformer-driven scaling of convnets for medical image segmentation.
\newblock In \emph{International Conference on Medical Image Computing and Computer-Assisted Intervention}, 405--415. Springer.

\bibitem[{Sadegheih et~al.(2024)Sadegheih, Bozorgpour, Kumari, Azad, and Merhof}]{sadegheih2024lhunet}
Sadegheih, Y.; Bozorgpour, A.; Kumari, P.; Azad, R.; and Merhof, D. 2024.
\newblock Lhu-net: A light hybrid u-net for cost-efficient, high-performance volumetric medical image segmentation.
\newblock \emph{arXiv preprint arXiv:2404.05102}.

\bibitem[{Sun et~al.(2023)Sun, Liu, Bair, and Kolter}]{sun2023wanda}
Sun, M.; Liu, Z.; Bair, A.; and Kolter, J.~Z. 2023.
\newblock A simple and effective pruning approach for large language models.
\newblock \emph{arXiv preprint arXiv:2306.11695}.

\bibitem[{Ulyanov, Vedaldi, and Lempitsky(2016)}]{ulyanov2016insnorm}
Ulyanov, D.; Vedaldi, A.; and Lempitsky, V. 2016.
\newblock Instance normalization: The missing ingredient for fast stylization.
\newblock \emph{arXiv preprint arXiv:1607.08022}.

\bibitem[{Wasserthal et~al.(2023)Wasserthal, Breit, Meyer, Pradella, Hinck, Sauter, Heye, Boll, Cyriac, Yang, Bach, and Segeroth}]{wasserthal2023totalsegmentator}
Wasserthal, J.; Breit, H.-C.; Meyer, M.~T.; Pradella, M.; Hinck, D.; Sauter, A.~W.; Heye, T.; Boll, D.~T.; Cyriac, J.; Yang, S.; Bach, M.; and Segeroth, M. 2023.
\newblock TotalSegmentator: Robust Segmentation of 104 Anatomic Structures in CT Images.
\newblock \emph{Radiology: Artificial Intelligence}, 5(5): e230024.

\bibitem[{Xie et~al.(2021)Xie, Zhang, Shen, and Xia}]{xie2021cotr}
Xie, Y.; Zhang, J.; Shen, C.; and Xia, Y. 2021.
\newblock Cotr: Efficiently bridging cnn and transformer for 3d medical image segmentation.
\newblock In \emph{Medical Image Computing and Computer Assisted Intervention--MICCAI 2021: 24th International Conference, Strasbourg, France, September 27--October 1, 2021, Proceedings, Part III 24}, 171--180. Springer.

\bibitem[{Xie et~al.(2024)Xie, Zhang, Xia, and Wu}]{xie2024unimiss+}
Xie, Y.; Zhang, J.; Xia, Y.; and Wu, Q. 2024.
\newblock UniMiSS+: Universal medical self-supervised learning from cross-dimensional unpaired data.
\newblock \emph{IEEE Transactions on Pattern Analysis and Machine Intelligence}.

\bibitem[{Yadav et~al.(2025)Yadav, Rizk, Chen, and KC}]{yadav2025mlru++}
Yadav, N.~K.; Rizk, R.; Chen, W.~C.; and KC. 2025.
\newblock MLRU++: Multiscale Lightweight Residual UNETR++ with Attention for Efficient 3D Medical Image Segmentation.

\bibitem[{Zhang et~al.(2021)Zhang, Xie, Xia, and Shen}]{zhang2021dodnet}
Zhang, J.; Xie, Y.; Xia, Y.; and Shen, C. 2021.
\newblock Dodnet: Learning to segment multi-organ and tumors from multiple partially labeled datasets.
\newblock In \emph{Proceedings of the IEEE/CVF conference on computer vision and pattern recognition}, 1195--1204.

\bibitem[{Zhou et~al.(2021)Zhou, Guo, Zhang, Yu, Wang, and Yu}]{zhou2021nnformer}
Zhou, H.-Y.; Guo, J.; Zhang, Y.; Yu, L.; Wang, L.; and Yu, Y. 2021.
\newblock nnformer: Interleaved transformer for volumetric segmentation.
\newblock \emph{arXiv preprint arXiv:2109.03201}.

\end{thebibliography}

\clearpage
\appendix

\section{Implementation Details}
\label{app:implementation}
\begin{figure}[h]
\centerline{\includegraphics[width=\linewidth]{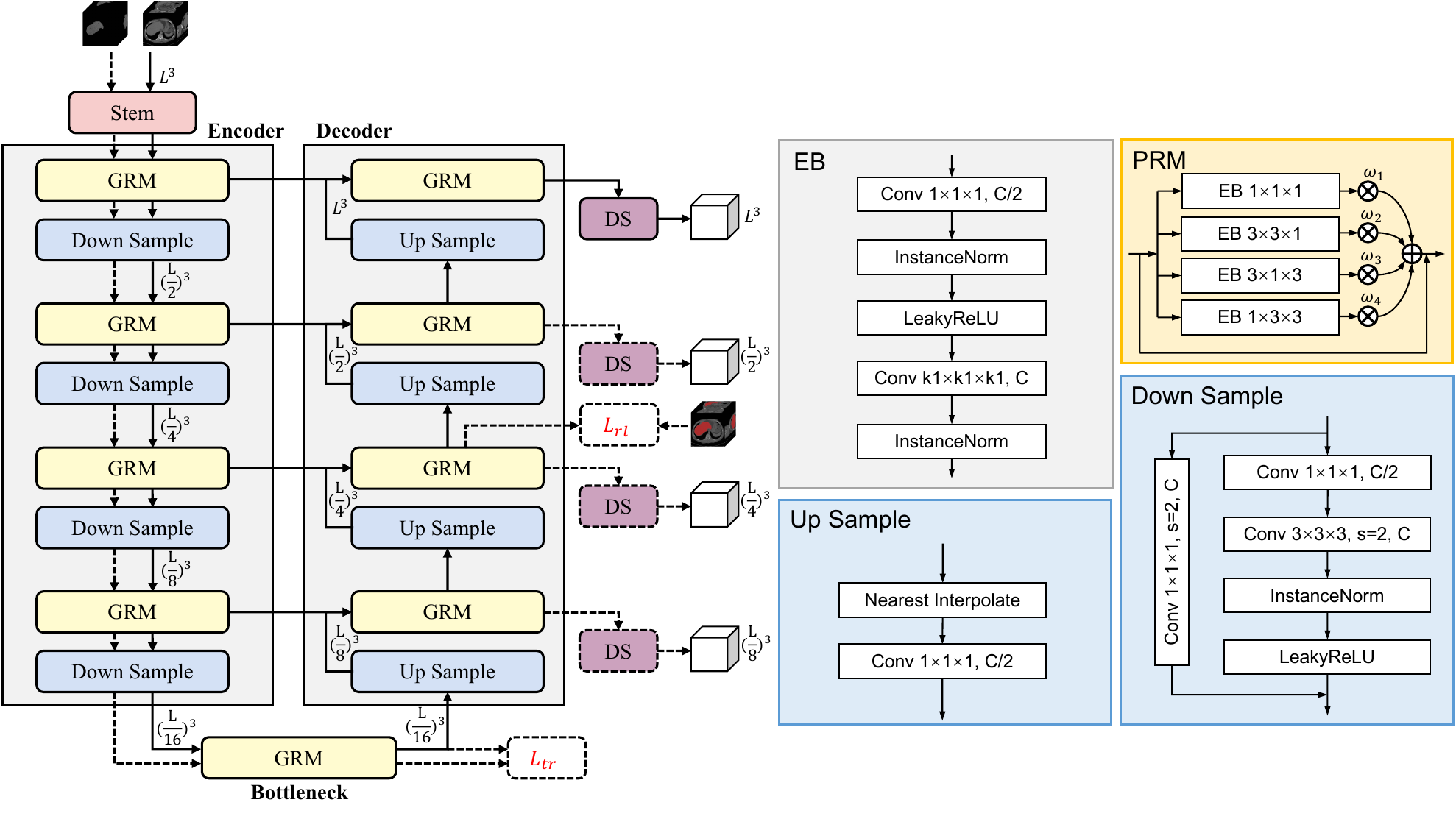}}
\caption{The detailed architecture for PSP-Seg-S model.}
\label{fig:model_arch}
\end{figure}

To achieve capacity-efficient segmentation, we scale the PSP-Seg models along multiple dimensions. PSP-Seg-S, the smallest variant, has a depth of 5 and channel dimensions of 16, 32, 64, 128, and 256 across the encoder. PRM employs 4 parallel branches using efficient block (EB) convolution kernels of sizes: $1 \times 1 \times 1$, $1 \times 3 \times 3$, $3 \times 1 \times 3$, and $3 \times 3 \times 1$.
PSP-Seg-B shares the same depth as PSP-Seg-S but uses 7 branches in PRM. The EB convolution kernels used are: $1 \times 1 \times 1$, $1 \times 1 \times 3$, $1 \times 3 \times 1$, $3 \times 1 \times 1$, $1 \times 3 \times 3$, $3 \times 1 \times 3$, and $3 \times 3 \times 1$.
The model parameters and resource consumption for each variant are summarized in Table \ref{tab:basemodel}. We visualize the detailed backbone structure of PSP-Seg-S in Figure \ref{fig:model_arch}.

\begin{table}[htbp]
\centering
{\small
\setlength{\tabcolsep}{0.6mm}
\begin{tabular}{cccccc}
\hline
Model     & Channels             & PN & GPU M. & Time & \#Params \\ \hline
PSP-Seg-S & 16,32,64,128,256     & 4   & 4.5          & 66                & 5.4          \\
PSP-Seg-B & 16,32,64,128,256     & 7   & 7.7          & 94                & 13.2         \\
PSP-Seg-L & 16,32,64,128,256,320 & 7   & 7.8          & 101               & 33.1         \\ \hline
\end{tabular}
}
\caption{The hyper-parameters for building the redundant base model variants. PN represents the number of the parallel efficient block in a PRM. GPU M. and Time denote the GPU memory usage (GB) and epoch time (s) during training. \#Params is model's parameter numbers (M).}
\label{tab:basemodel}
\end{table}

\section{Convergence and Improvement Checks in Progressive Pruning}
Progressive Pruning is primarily guided by the FD loss, which informs both pruning and restoration decisions. Below, we describe the procedures for the convergence check and improvement check, both of which are based on the FD loss.

The convergence check is conducted at the beginning of each epoch to determine whether the model has reached a stable state. Specifically, we examine the historical FD loss and deem the model to have converged if neither the TR loss nor the RL loss has improved over the past 10 epochs. This assessment is restricted to epochs occurring after the most recent pruning operation.

The improvement check evaluates whether further pruning is warranted or whether partial restoration is necessary. A naive approach would directly compare the current FD loss with its historical minimum; however, such comparisons can be sensitive to random fluctuations. To mitigate this, we define a performance maintenance threshold, the historical best FD loss plus 0.01. If the current FD loss exceeds this threshold, we consider the model to be over-pruned.

\begin{algorithm}[h]
\caption{Block-wise Pruning}
\label{alg:blockwise_prune}
\textbf{Input}: Model $f_\theta$, calibration dataset $\mathcal{D}_c$ \\
\textbf{Parameter}: Pruning step size $p$ \\
\textbf{Output}: Pruned model $f'_\theta$, masked branch set $\mathcal{B}_{\text{masked}}$
\begin{algorithmic}[1] %[1] enables line numbers
\STATE Initialize $\mathcal{B}_{\text{masked}} \gets \emptyset$
\STATE Run inference on $\mathcal{D}_c$ and cache intermediate outputs $\{(x_l, y_l)\}$ for each prunable module $PRM_l$
\FOR{each prunable module $PRM_l$ in $f_\theta$}
    \STATE $\mathcal{B}_l \gets$ the set of all branches in $PRM_l$
    \FOR{each branch subset $\mathcal{P} \subseteq \mathcal{B}_l$ with cardinality $p$}
        \STATE $PRM_l^{(\mathcal{P})} \gets$ Apply mask $\mathcal{P}$ to $PRM_l$
        \STATE Compute output: $\hat{y}_l \gets PRM_l^{(\mathcal{P})}(x_l)$
        \STATE Compute discrepancy: $d_{\mathcal{P}} \gets \left\| \hat{y}_l - y_l \right\|_F$
        \STATE Store $(\mathcal{P}, d_{\mathcal{P}})$ in discrepancy dictionary $\mathcal{D}_l$
    \ENDFOR
    \STATE Select optimal subset: $\mathcal{P}^* \gets \arg\min_{\mathcal{P}} \mathcal{D}_l[\mathcal{P}]$
    \STATE Update masked branch set: $\mathcal{B}_{\text{masked}} \gets \mathcal{B}_{\text{masked}} \cup \{(PRM_l, \mathcal{P}^*)\}$
    \STATE Replace $PRM_l$ with $PRM_l^{(\mathcal{P}^*)}$ in $f_\theta$
\ENDFOR
\STATE \textbf{return} Pruned model $f'_\theta$, masked branch set $\mathcal{B}_{\text{masked}}$
\end{algorithmic}
\end{algorithm}

\section{Details of Datasets}
\label{app:datasets}
We conducted experiments on five publicly available datasets: LiTS, KiTS, BraTS21, Hepatic Vessel, and Colon. The LiTS dataset contains 131 contrast-enhanced 3D abdominal CT scans with liver and tumor annotations. The KiTS dataset comprises CT scans from kidney cancer patients who underwent nephrectomy, annotated for kidneys and kidney tumors. The BraTS21 dataset provides annotations for brain tumors across four MRI modalities (T1, T1-weighted, T2-weighted, and T2-FLAIR), providing segmentation for all 1,251 paired multi-modal images on peritumoral edematous/invaded tissue, the necrotic tumor core, and the Gd-enhancing tumor. The Hepatic Vessel and Colon datasets are part of the Medical Segmentation Decathlon (MSD) challenge  \cite{antonelli2022medical}. The Hepatic Vessel dataset includes 303 liver CT scans annotated for hepatic vessels and tumors, while the Colon dataset contains 126 contrast-enhanced CT scans focused on primary colon cancer segmentation. All datasets were randomly split into training and testing sets using an 80:20 ratio, following the same protocol as in DoDNet \cite{zhang2021dodnet}.

\begin{table*}[htbp]
\centering
{\small
\setlength{\tabcolsep}{1.2mm}
\begin{tabular}{cccccccccc|ccccccccc}
\hline
\multirow{3}{*}{Model} & \multicolumn{9}{c|}{LiTS}                                     & \multicolumn{9}{c}{KiTS}                                     \\ \cline{2-19} 
 &
  \multicolumn{2}{c}{GPU M.} &
  \multicolumn{2}{c}{Time} &
  \multirow{2}{*}{\#P} &
  \multicolumn{2}{c}{Liver} &
  \multicolumn{2}{c|}{Tumor} &
  \multicolumn{2}{c}{GPU M.} &
  \multicolumn{2}{c}{Time} &
  \multirow{2}{*}{\#P} &
  \multicolumn{2}{c}{Kidney} &
  \multicolumn{2}{c}{Tumor} \\ \cline{2-5} \cline{7-14} \cline{16-19} 
                       & Tr.  & Inf. & Tr. & Inf. &       & DSC  & NSD  & DSC  & NSD  & Tr.  & Inf. & Tr. & Inf. &       & DSC  & NSD  & DSC  & NSD  \\ \hline
nnU-Net                 & 6.6  & 3.4  & 86  & 153  & 31.2  & 95.8 & 88.7 & 64.5 & 69.8 & 6.6  & 3.8  & 86  & 174  & 31.2  & 96.7 & 96.1 & 84.8 & 81.7 \\
CoTr                   & 7.7  & 3.4  & 173 & 247  & 41.9  & 94.0 & 85.0 & 64.2 & 65.5 & 7.7  & 3.8  & 173 & 277  & 41.9  & 96.3 & 95.0 & 81.3 & 78.9 \\
nnFormer               & 7.4  & 4.1  & 65  & 128  & 151.0 & 95.8 & 88.4 & 64.8 & 69.0 & 7.4  & 4.6  & 65  & 136  & 151.0 & 96.6 & 95.8 & 80.6 & 77.9 \\
SwinUNETR-V2              & 21.4 & 4.0  & 147 & 240  & 72.8  & 95.5 & 87.2 & 60.7 & 64.4 & 21.4 & 4.4  & 147 & 271  & 72.8  & 96.7 & 96.0 & 81.6 & 78.9 \\
U-Mamba                & 11.8 & 3.9  & 130 & 233  & 70.0  & 95.6 & 87.6 & 65.2 & 69.9 & 11.8 & 4.3  & 130 & 275  & 70.0  & 96.6 & 95.9 & 84.3 & 80.8 \\
STU-Net*               & 8.0  & 3.7  & 88  & 172  & 58.2  & 96.3 & 89.9 & 65.5 & 69.1 & 8.0  & 4.1  & 88  & 187  & 58.2  & 96.6 & 95.8 & 85.4 & 80.8 \\
UniMiSS+*              & 6.3  & 3.1  & 61  & 102  & 54.2  & 96.0 & 89.1 & 65.5 & 71.0 & 6.3  & 3.5  & 61  & 126  & 54.2  & 96.6 & 95.9 & 84.2 & 81.7 \\ \hline
DMF-Net                & 3.6  & 2.5  & 36  & 57   & 3.9   & 95.3 & 87.3 & 61.0 & 64.6 & 3.6  & 3.0  & 36  & 59   & 3.9   & 96.0 & 94.9 & 71.7 & 67.9 \\
SegFormer3D            & 2.1  & 2.2  & 26  & 36   & 4.5   & 95.6 & 87.1 & 60.9 & 65.2 & 2.1  & 2.7  & 26  & 34   & 4.5   & 96.3 & 95.2 & 75.4 & 71.9 \\
LHU-Net                & 5.3  & 2.8  & 232 & 200  & 13.0  & 95.1 & 86.1 & 60.5 & 65.7 & 5.3  & 3.2  & 232 & 221  & 13.0  & 96.4 & 95.4 & 78.2 & 75.2 \\
Slim   UNETR           & 1.2  & 2.1  & 34  & 86   & 1.8   & 84.9 & 61.1 & 25.1 & 19.3 & 1.2  & 2.5  & 34  & 86   & 1.8   & 95.3 & 92.8 & 57.5 & 48.6 \\
EffiDec3D              & 3.7  & 2.5  & 126 & 129  & 3.2   & 94.8 & 85.8 & 61.9 & 66.4 & 3.7  & 2.9  & 126 & 172  & 3.2   & 95.6 & 94.9 & 78.9 & 73.6 \\
MLRU++                 & 3.8  & 2.8  & 61  & 120  & 46.1  & 94.7 & 85.6 & 63.3 & 66.6 & 3.8  & 3.3  & 61  & 133  & 46.1  & 96.2 & 94.9 & 71.4 & 66.9 \\
PSP-Seg-S*               & 3.8  & 2.5  & 52  & 60   & 4.3   & 95.2 & 85.8 & 63.1 & 67.1 & 3.7  & 2.9  & 47  & 74   & 4.2   & 96.3* & 95.3 & 81.1 & 75.7 \\
PSP-Seg-B*               & 5.0  & 2.7  & 62  & 111  & 10.3  & 95.5 & 86.3 & 65.0 & 68.4 & 5.5  & 3.1  & 74  & 171  & 10.7  & 96.6* & 95.9* & 82.0 & 79.1 \\
PSP-Seg-L*               & 5.2  & 2.6  & 63  & 148  & 19.8  & 96.1* & 87.3 & 65.9* & 69.6 & 6.3  & 3.1  & 90  & 180  & 23.4  & 96.6* & 95.8* & 84.5* & 80.5 \\ \hline
\end{tabular}
}
\caption{Resource consumption and detailed segmentation performance on the LiTS and KiTS datasets.}
\label{tab:lits&kits}
\end{table*}

\begin{table*}[htbp]
\centering
{\small
\setlength{\tabcolsep}{0.8mm}
\begin{tabular}{cccccccccccc|ccccccccc}
\hline
\multirow{3}{*}{Model} & \multicolumn{11}{c|}{BraTS21}                                                    & \multicolumn{9}{c}{HepaticVessel}                             \\ \cline{2-21} 
 &
  \multicolumn{2}{c}{GPU M.} &
  \multicolumn{2}{c}{Time} &
  \multirow{2}{*}{\#P} &
  \multicolumn{2}{c}{ED} &
  \multicolumn{2}{c}{NE} &
  \multicolumn{2}{c|}{EN} &
  \multicolumn{2}{c}{GPU M.} &
  \multicolumn{2}{c}{Time} &
  \multirow{2}{*}{\#P} &
  \multicolumn{2}{c}{HV} &
  \multicolumn{2}{c}{Tumor} \\ \cline{2-5} \cline{7-16} \cline{18-21} 
                       & Tr.  & Inf. & Tr. & Inf. &       & DSC   & NSD   & DSC   & NSD   & DSC   & NSD   & Tr.  & Inf. & Tr. & Inf. &       & DSC  & NSD   & DSC  & NSD  \\ \hline
nnU-Net                & 6.6  & 1.6  & 86  & 3    & 31.2  & 88.4  & 93.4  & 78.9  & 84.7  & 87.8  & 93.3  & 6.6  & 2.2  & 86  & 47   & 31.2  & 63.8 & 79.1  & 74.4 & 66.6 \\
CoTr                   & 7.7  & 1.7  & 173 & 5    & 41.9  & 86.9  & 92.7  & 76.3  & 82.4  & 87.1  & 93.4  & 7.7  & 2.2  & 173 & 66   & 41.9  & 61.7 & 75.7  & 62.0 & 48.4 \\
nnFormer               & 7.4  & 2.4  & 65  & 3    & 151.0 & 87.7  & 93.1  & 77.1  & 83.5  & 88.3  & 94.4  & 7.4  & 2.9  & 65  & 48   & 151.0 & 65.4 & 80.1  & 68.9 & 60.2 \\
SwinUNETR-V2           & 21.4 & 1.8  & 147 & 5    & 72.8  & 87.4  & 92.4  & 76.8  & 82.7  & 87.2  & 93.0  & 21.4 & 2.8  & 147 & 71   & 72.8  & 65.9 & 80.0  & 67.6 & 56.3 \\
U-Mamba                & 11.8 & 1.8  & 130 & 5    & 70.0  & 88.3  & 93.2  & 78.4  & 84.1  & 87.5  & 92.7  & 11.8 & 2.7  & 130 & 70   & 70.0  & 67.5 & 81.1  & 71.8 & 63.1 \\
STU-Net*               & 8.0  & 1.9  & 88  & 4    & 58.2  & 88.1  & 93.3  & 78.7  & 84.7  & 87.8  & 93.3  & 8.0  & 2.5  & 88  & 55   & 58.2  & 66.7 & 81.1  & 72.3 & 63.9 \\
UniMiSS+*              & 6.3  & 1.5  & 61  & 3    & 54.2  & 87.8  & 93.1  & 77.3  & 83.4  & 87.4  & 93.2  & 6.3  & 1.9  & 61  & 30   & 54.2  & 66.4 & 80.9  & 72.7 & 63.7 \\ \hline
DMF-Net                & 3.6  & 0.8  & 36  & 1    & 3.9   & 87.4  & 92.8  & 76.4  & 82.6  & 87.7  & 94.0  & 3.6  & 1.4  & 36  & 19   & 3.9   & 63.9 & 78.2  & 65.2 & 56.5 \\
SegFormer3D            & 2.1  & 0.6  & 26  & 1    & 4.5   & 85.7  & 92.4  & 71.2  & 80.5  & 84.3  & 92.8  & 2.1  & 1.1  & 26  & 19   & 4.5   & 62.6 & 78.0  & 66.7 & 57.0 \\
LHU-Net                & 5.3  & 1.0  & 232 & 4    & 13.0  & 87.2  & 92.7  & 78.2  & 83.9  & 86.7  & 92.2  & 5.3  & 1.7  & 232 & 72   & 13.0  & 63.9 & 78.8  & 67.8 & 59.0 \\
SlimUNETR              & 1.2  & 0.5  & 34  & 2    & 1.8   & 77.5  & 85.5  & 61.6  & 70.9  & 76.3  & 86.6  & 1.2  & 1.0  & 34  & 22   & 1.8   & 47.6 & 63.1  & 30.8 & 13.5 \\
EffiDec3D              & 3.7  & 0.8  & 126 & 3    & 3.2   & 84.1  & 92.4  & 72.6  & 82.9  & 82.7  & 92.8  & 3.7  & 1.2  & 126 & 37   & 3.2   & 63.1 & 78.9  & 68.1 & 59.3 \\
MLRU++                 & 3.8  & 1.0  & 61  & 3    & 46.1  & 87.7  & 93.3  & 77.6  & 83.3  & 87.3  & 93.4  & 3.8  & 1.7  & 61  & 45   & 46.1  & 65.5 & 80.1  & 68.5 & 59.4 \\
PSP-Seg-S*               & 3.7  & 0.8  & 61  & 2    & 5.1   & 87.6  & 93.0* & 77.4  & 83.1  & 87.9  & 93.7* & 3.6  & 1.4  & 45  & 20   & 5.2   & 65.1 & 79.9  & 72.9 & 63.5 \\
PSP-Seg-B*               & 5.2  & 0.9  & 68  & 3    & 11.7  & 87.9* & 93.3* & 77.9  & 83.6  & 87.7  & 93.5  & 5.0  & 1.5  & 64  & 35   & 9.6   & 65.4 & 80.2  & 71.2 & 62.6 \\
PSP-Seg-L*               & 5.3  & 1.0  & 70  & 3    & 24.2  & 88.1* & 93.4* & 78.9* & 84.6* & 88.5* & 94.3* & 5.2  & 1.6  & 68  & 38   & 22.0  & 66.5 & 81.5* & 73.1 & 63.5 \\ \hline
\end{tabular}
}
\caption{Resource consumption and detailed segmentation performance on the BraTS21 and HepaticVessel datasets.}
\label{tab:brats&hv}
\end{table*}

\begin{table*}[htbp]
\centering
{\small
\setlength{\tabcolsep}{1.5mm}
\begin{tabular}{cccccccccccccccc}
\hline
\multirow{2}{*}{Model} &
  \multirow{2}{*}{PRM} &
  \multirow{2}{*}{FD Loss} &
  \multirow{2}{*}{PSP} &
  \multirow{2}{*}{PT} &
  \multicolumn{2}{c}{GPU Mem.} &
  \multicolumn{2}{c}{Time} &
  \multirow{2}{*}{\#P} &
  \multicolumn{2}{c}{Liver} &
  \multicolumn{2}{c}{Tumor} &
  \multicolumn{2}{c}{AVG} \\ \cline{6-9} \cline{11-16} 
                   &   &   &   &   & Tr. & Inf. & Tr. & Inf. &      & DSC           & NSD           & DSC           & NSD           & DSC           & NSD           \\ \hline
\multirow{5}{*}{S} & \ding{55} & \ding{55} & \ding{55} & \ding{55} & \textbf{3.6} & \textbf{2.5}  & \textbf{38}  & \textbf{51}   & \textbf{3.7}  & 94.2          & 83.9          & 58.3          & 61.6          & 76.3          & 72.8          \\
                   & \ding{51} & \ding{55} & \ding{55} & \ding{55} & 4.5 & 2.6  & 65  & 122  & 5.4  & 95.0          & \textbf{86.2} & 62.3          & 66.5          & 78.6          & 76.3          \\
                   & \ding{51} & \ding{51} & \ding{55} & \ding{55} & 4.5 & 2.7  & 66  & 124  & 5.4  & 94.8          & 85.3          & 62.5          & 65.8          & 78.7          & 75.6          \\
                   & \ding{51} & \ding{51} & \ding{51} & \ding{55} & 4.5 & 2.6  & 65  & 122  & 5.4  & 94.9          & 85.2          & 61.8          & 65.2          & 78.3          & 75.2          \\
                   & \ding{51} & \ding{51} & \ding{51} & \ding{51} & 3.8 & \textbf{2.5}  & 52  & 60   & 4.3  & \textbf{95.2} & 85.8          & \textbf{63.1} & \textbf{67.1} & \textbf{79.1} & \textbf{76.4} \\ \hline
\multirow{5}{*}{B} & \ding{55} & \ding{55} & \ding{55} & \ding{55} & \textbf{4.5} & \textbf{2.6}  & \textbf{39}  & \textbf{107}  & \textbf{4.1}  & 94.5          & 85.0          & 59.9          & 63.8          & 77.2          & 74.4          \\
                   & \ding{51} & \ding{55} & \ding{55} & \ding{55} & 7.7 & 2.8  & 93  & 171  & 13.2 & 95.1          & 86.2          & 63.7          & 67.7          & 79.4          & 77.0          \\
                   & \ding{51} & \ding{51} & \ding{55} & \ding{55} & 7.7 & 2.8  & 94  & 173  & 13.2 & 95.1          & 85.7          & 63.6          & 67.8          & 79.4          & 76.7          \\
                   & \ding{51} & \ding{51} & \ding{51} & \ding{55} & 5.0 & 2.7  & 73  & 127  & 9.1  & 95.1          & 85.9          & 63.1          & 67.0          & 79.1          & 76.5          \\
                   & \ding{51} & \ding{51} & \ding{51} & \ding{51} & 5.0 & 2.7  & 62  & 111  & 10.3 & \textbf{95.5} & \textbf{86.3} & \textbf{65.0} & \textbf{68.4} & \textbf{80.3} & \textbf{77.5} \\ \hline
\end{tabular}
}
\caption{Ablation study of the PRM, FD Loss, PSP, and pre-training components in PSP-Seg on the LiTS dataset. PRM refers to the Parallel Redundant Module, FD Loss denotes the Functional Decoupling Loss, PSP stands for Progressive Pruning, and PT indicates the use of pre-trained weights. The best performance for each metric is highlighted in \textbf{bold}.}
\label{tab:ablation_app}
\end{table*}

\begin{table}[htbp]
\centering
{\small
\setlength{\tabcolsep}{0.8mm}
\begin{tabular}{cccccccccccc}
\hline
\multirow{2}{*}{P} &
  \multicolumn{2}{c}{GPU M.} &
  \multicolumn{2}{c}{Time} &
  \multirow{2}{*}{\#P} &
  \multicolumn{2}{c}{Liver} &
  \multicolumn{2}{c}{Tumor} &
  \multicolumn{2}{c}{AVG} \\ \cline{2-5} \cline{7-12} 
  & Tr. & Inf. & Tr. & Inf. &      & DSC           & NSD           & DSC           & NSD           & DSC           & NSD           \\ \hline
1 & 6.4 & 2.7  & 74  & 173  & 27.1 & 95.7          & \textbf{88.0} & 65.1          & 69.1          & 80.4          & \textbf{78.6} \\
2 & \textbf{5.2} & \textbf{2.6}  & \textbf{63}  & \textbf{148}  & \textbf{19.8} & \textbf{96.1} & 87.3          & \textbf{65.9} & \textbf{69.6} & \textbf{81.0} & 78.5          \\
3 & 5.7 & \textbf{2.6}  & 69  & 161  & 22.8 & 95.6          & 87.3          & 63.6          & 67.3          & 79.6          & 77.3          \\
4 & 6.4 & 2.7  & 75  & 172  & 27.1 & 95.4          & 87.2          & 63.7          & 67.1          & 79.2          & 77.2          \\ \hline
\end{tabular}
}
\caption{Experiments on four initial prune steps.}
\label{tab:prune_step}
\end{table}

\begin{table}[htbp]
\centering
{\small
\setlength{\tabcolsep}{0.8mm}
\begin{tabular}{cccccccccccc}
\hline
\multirow{2}{*}{CS} & \multicolumn{2}{c}{GPU M.}  & \multicolumn{2}{c}{Time}   & \multirow{2}{*}{\#P} & \multicolumn{2}{c}{Liver}     & \multicolumn{2}{c}{Tumor}     & \multicolumn{2}{c}{AVG}       \\ \cline{2-5} \cline{7-12} 
                    & Tr.          & Inf.         & Tr.         & Inf.         &                      & DSC           & NSD           & DSC           & NSD           & DSC           & NSD           \\ \hline
50                  & 6.4          & 2.7          & 75          & 172          & 27.1                 & 94.6          & \textbf{87.7} & 63.5          & 66.8          & 79.1          & 77.3          \\
100                 & \textbf{5.2} & \textbf{2.6} & \textbf{63} & \textbf{148} & \textbf{19.8}        & \textbf{96.1} & 87.3          & \textbf{65.9} & \textbf{69.6} & \textbf{81.0} & \textbf{78.5} \\ \hline
\end{tabular}
}
\caption{Experiments on two calibration set sizes. CS denotes the calibration set.}
\label{tab:calib_size}
\end{table}

\begin{table}[htbp]
\centering
{\small
\setlength{\tabcolsep}{0.8mm}
\begin{tabular}{ccccccccc}
\hline
\multirow{2}{*}{Model} & \multirow{2}{*}{Pre-train} & \multirow{2}{*}{Method} & \multicolumn{2}{c}{Liver} & \multicolumn{2}{c}{Tumor} & \multicolumn{2}{c}{AVG}       \\ \cline{4-9} 
                   &                    &         & DSC           & NSD           & DSC           & NSD           & DSC           & NSD           \\ \hline
\multirow{4}{*}{S}     & \multirow{2}{*}{\ding{55}}        & Re-train                 & 94.7        & 85.1        & 61.8    & \textbf{65.6}   & \textbf{78.3} & \textbf{75.3} \\
                   &                    & PSP & \textbf{94.9} & \textbf{85.2} & \textbf{61.8} & 65.2          & \textbf{78.3} & 75.2          \\ \cline{2-9} 
                   & \multirow{2}{*}{\ding{51}} & Re-train & 95.1          & \textbf{86.4} & 62.9          & 66.4          & 79.0          & \textbf{76.4} \\
                   &                    & PSP & \textbf{95.2} & 85.8          & \textbf{63.1} & \textbf{67.1} & \textbf{79.1} & \textbf{76.4} \\ \hline
\multirow{4}{*}{B} & \multirow{2}{*}{\ding{55}} & Re-train & 94.7          & 85.4          & 60.6          & 64.0          & 77.6          & 74.7          \\
                   &                    & PSP & \textbf{95.1} & \textbf{85.9} & \textbf{63.1} & \textbf{67.0} & \textbf{79.1} & \textbf{76.5} \\ \cline{2-9} 
                   & \multirow{2}{*}{\ding{51}} & Re-train & 95.3          & 86.1          & \textbf{65.1} & 68.2          & 80.2          & 77.2          \\
                   &                    & PSP & \textbf{95.5} & \textbf{86.3} & 65.0          & \textbf{68.4} & \textbf{80.3} & \textbf{77.5} \\ \hline
\end{tabular}
}
\caption{Segmentation performance of the re-trained model and the model generated by PSP. The best performance is highlighted in \textbf{bold}. Our PSP-Seg model outperforms all retrained baseline models.}
\label{tab:retrain_app}
\end{table}

\begin{figure}[t]
\centerline{\includegraphics[width=0.95\linewidth]{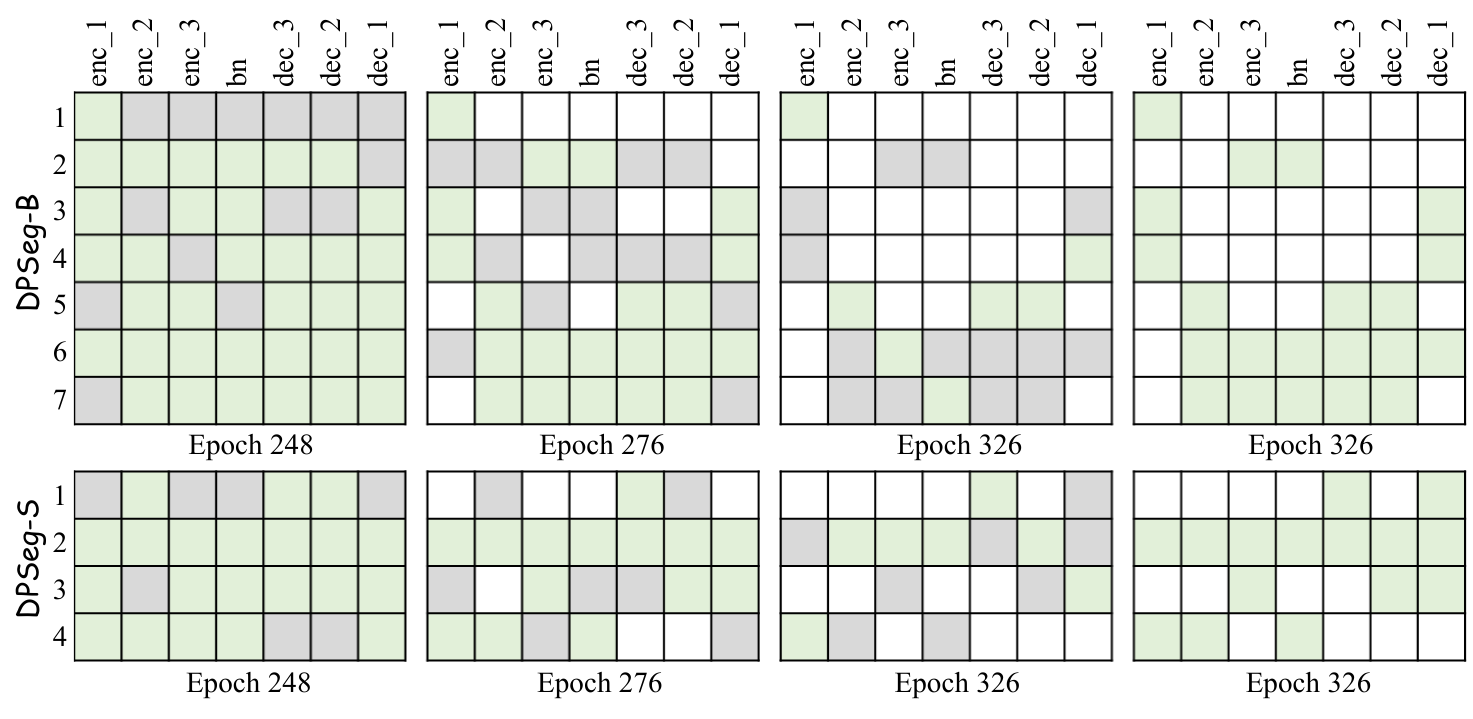}}
\caption{Additional visualization for PSP-Seg on LiTS dataset. We provide the visualization on different PSP-Seg variants. For PSP-Seg-S, the labels `1–4' denote efficient blocks with kernel sizes of $1\times1\times1$, $1\times3\times3$, $3\times1\times3$, and $3\times3\times1$, respectively.}
\label{fig:prune_progress_lits}
\end{figure}

\begin{figure*}[t]
\centerline{\includegraphics[width=0.95\textwidth]{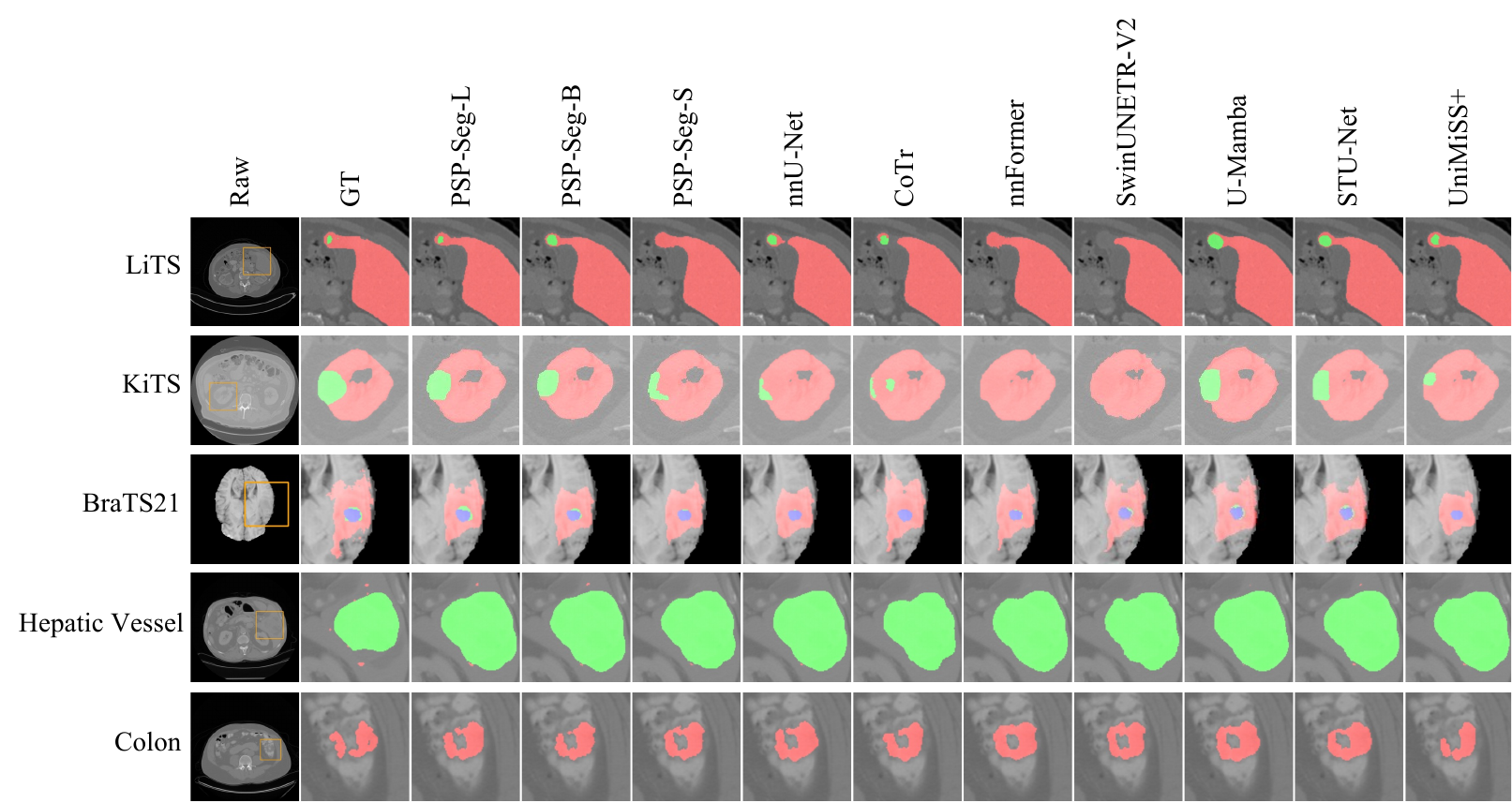}}
\caption{Visualization of the segmentation results comparing PSP-Seg with advance models. We highlight organs in red and tumors in green. For BraTS21, red is used to indicate the edema region, green denotes the non-enhancing tumor core, and blue represents the enhancing tumor area.}
\label{fig:seg_visual_advance}
\end{figure*}

\begin{figure*}[t]
\centerline{\includegraphics[width=0.95\textwidth]{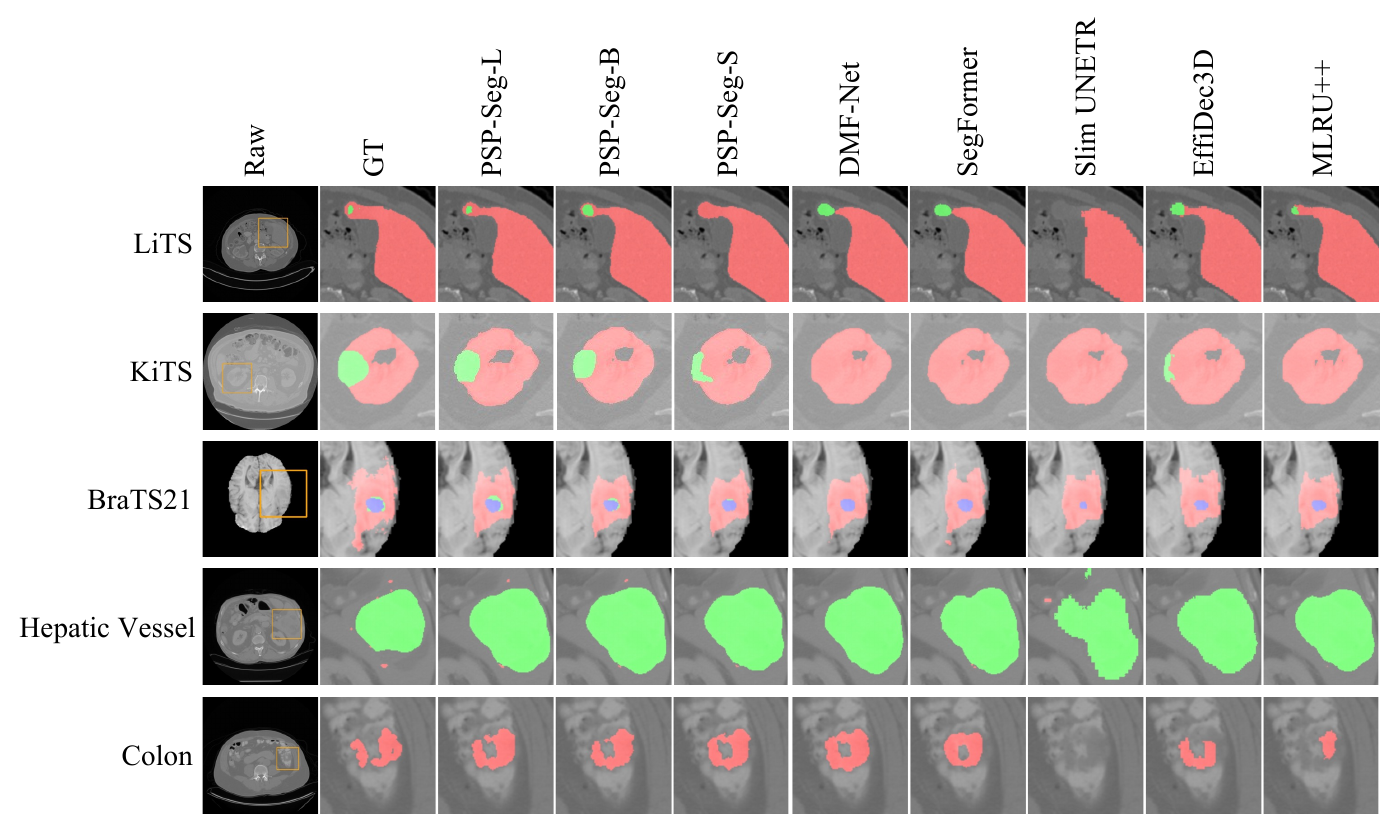}}
\caption{Visualization of the segmentation results comparing PSP-Seg with efficient models. We highlight organs in red and tumors in green. For BraTS21, red is used to indicate the edema region, green denotes the non-enhancing tumor core, and blue represents the enhancing tumor area.}
\label{fig:seg_visual_efficient}
\end{figure*}

\section{Comparing Block-wise Pruning with Lu \textit{et al.} \cite{lu2024not}}
Lu \textit{et al.} proposed a post-training pruning strategy aimed at effectively removing non-contributing experts from a Mixture-of-Experts (MoE) model. To identify the least impactful experts, they introduced a heuristic search method that enumerates all possible pruning combinations within each MoE layer and computes the discrepancy between the original and pruned layers to determine the optimal pruning configuration. 
Although our Block-wise Pruning strategy shares some surface-level similarities with Lu \textit{et al.}'s approach, it differs substantially in terms of purpose, methodology, and implementation.

Purpose: Our Block-wise Pruning is an integral part of a progressive pruning framework designed for pruning during training. In contrast, Lu \textit{et al.}'s method is applied post-training.

Methodology: Methodologically, Block-wise Pruning adopts a mask-then-prune strategy within the progressive pruning pipeline, which helps mitigate the risk of over-pruning.  Lu \textit{et al.}'s approach, however, does not explicitly address over-pruning and relies on manually tuned hyper-parameters to set the pruning ratio.

Implementation: From an implementation perspective, Block-wise Pruning operates on a randomly sampled patch dataset, offering computational efficiency and adaptability to various data distributions. In contrast,  Lu \textit{et al.}'s method requires access to large-scale pre-training datasets such as C4, resulting in higher computational and storage demands and limiting its applicability in diverse domains.

\section{Pseudo Code of Block-wise Pruning}
To enhance understanding of the proposed Block-wise Pruning strategy, we present the corresponding pseudo code in Algorithm \ref{alg:blockwise_prune}. For each prunable PRM module in the model, the algorithm systematically evaluates different combinations of branch masks and selects the subset that minimally affects the model output. This selective masking facilitates efficient model compression while preserving performance.

\section{Experiments Results}

\subsection{Detailed Results of Segmentation Performance}
\label{app:Main exp}
We further reported more details of the proposed results in Section 4.4. Note that the resource consumption is calculated on an NVIDIA 2080 Ti GPU for most models, except for SwinUNETR-V2 and U-Mamba, which are calculated using an NVIDIA 3090 due to their high GPU memory requirements. 

Shown in Table \ref{tab:lits&kits} and Table \ref{tab:brats&hv}, we reported the individual results of each foreground category. These results further reveal that (1) the performance gap between efficient and advanced models is mainly from tumor segmentation, which is a more challenging task than anatomical structures segmentation, such as the liver and kidney. (2) In contrast, our PSP-Seg achieves superior performance on tumor segmentation, being comparable to the results from the best models.

\subsection{Impact of Prune Step and Calibration Set Size}
We analyze two key factors in the proposed Progressive Pruning framework: the prune step size $p$ and the calibration set size $CS$. All experiments in this section are conducted using PSP-Seg-L on the LiTS dataset.

To investigate the effect of the prune step, we evaluated $p \in {1, 2, 3, 4}$, with results presented in Table \ref{tab:prune_step}. We find that $p = 2$ achieves the best overall balance between resource efficiency and segmentation performance. When $p$ is reduced to 1, resource consumption increases across all evaluation metrics, albeit with a marginal improvement in average NSD. This is attributed to more conservative pruning, resulting in slower architectural reduction and reduced tolerance to suboptimal pruning decisions (as $p$ approaches zero). Conversely, larger values of $p$ carry a higher risk of overly aggressive or unsuitable pruning.

We also examined the effect of calibration set size by testing $CS \in {50, 100}$, as shown in Table \ref{tab:calib_size}.  The results align with the expectation that a larger calibration set enhances pruning robustness. A smaller $CS$ makes the pruning process more sensitive to noise and variability, potentially leading to suboptimal model structures.

Based on these findings, we set $p = 2$ and $CS = 100$ in all subsequent experiments to fully exploit the capabilities of PSP-Seg.

\subsection{Ablation Studies of PSP-Seg-S and PSP-Seg-B}
\label{app:ablation}
We further performed ablation studies on the two remaining PSP-Seg variants, PSP-Seg-S and PSP-Seg-B, using the LiTS dataset, as summarized in Table  \ref{tab:ablation_app}. Consistent with earlier findings, the baseline model achieves the lowest computational cost but suffers from suboptimal segmentation performance. Incorporating the PRM module notably enhances segmentation accuracy, albeit at the expense of increased resource usage. In contrast, our proposed PSP-Seg models, when initialized with pre-trained weights, effectively balance computational efficiency and performance. Notably, they achieve substantial improvements in tumor extraction accuracy, even outperforming their redundant counterparts.

\section{Some Discussions}
\subsection{Re-training Results for PSP-Seg-S and PSP-Seg-B}
\label{app:retrain}
We present additional results for PSP-Seg-S and PSP-Seg-B under the re-training setting described in Section 5.1. As shown in Table \ref{tab:retrain_app}, unlike the behavior observed for PSP-Seg-L, the two smaller variants exhibit similar performance with and without pre-training.
This observation is consistent with the intuition that smaller models tend to prune fewer modules, thereby retaining most of the original structure and integrating only limited additional knowledge from the pruned components.

\subsection{Visualization of Progressive Pruning}
\label{app:pruneprogress}
We further extended our analysis by visualizing the pruning behavior of additional model variants (PSP-Seg-S and PSP-Seg-B) on the LiTS, as shown in Fig. \ref{fig:prune_progress_lits}.  

One key observation is that the extent and pattern of pruning vary, indicating that PSP-Seg is capable of generating task-specific models by adapting to the underlying characteristics. This adaptability supports our claim that PSP-Seg can flexibly tailor the network architecture to best suit the task at hand.

Moreover, we consistently observed that the blocks retained in the shallow layers (e.g., early encoder and late decoder stages) tend to have smaller kernel sizes, while the blocks preserved in the deeper layers (central parts of the model) more frequently employ larger kernels.

\subsection{Visualization of Segmentation Results}
We present qualitative comparisons of segmentation results from nnU-Net, CoTr, nnFormer, SwinUNETR-V2, U-Mamba, STU-Net, and UniMiSS+, and our PSP-Seg series on five datasets, as shown in Fig. \ref{fig:seg_visual_advance}. The visualizations demonstrate that the PSP-Seg series produces segmentations that closely match the ground truths (GTs), effectively mitigating both over-segmentation and under-segmentation. In addition, we compare our PSP-Seg series with several efficient models, including DMF-Net, SegFormer3D, Slim UNETR, EffiDec3D, MLRU++, and our PSP-Seg series, as shown in Fig. \ref{fig:seg_visual_efficient}. The superiority of PSP-Seg becomes more apparent in these comparisons, particularly in terms of segmentation accuracy and boundary precision.

\end{document}